\documentclass[sigconf]{acmart}

\usepackage{graphicx}
\graphicspath{{./images/}}
\usepackage{booktabs} 
\usepackage{multirow}
\usepackage{amsmath}  
\usepackage{subcaption}
\usepackage{caption}
\usepackage{multirow}
\usepackage{multicol}
\usepackage{hyperref}
\usepackage{url}
\usepackage{balance}
\usepackage{verbatim}
\usepackage[ruled,vlined,linesnumbered]{algorithm2e}
\usepackage{mathtools}

\newcommand{\prince}{\textsc{Prince}\xspace}

\theoremstyle{definition}

\DeclareMathOperator*{\argmax}{arg\,max}
\DeclareMathOperator*{\argmin}{arg\,min}

\setcopyright{acmcopyright}

\copyrightyear{2020}
\acmYear{2020}
\setcopyright{acmcopyright}
\acmConference[WSDM '20]{The Thirteenth ACM International Conference on Web Search and
	Data Mining}{February 3--7, 2020}{Houston, TX, USA}
\acmBooktitle{The Thirteenth ACM International Conference on Web Search and Data Mining
	(WSDM '20), February 3--7, 2020, Houston, TX, USA}
\acmPrice{15.00}

\begin{document}
	
\fancyhead{}

\title{PRINCE: Provider-side Interpretability with\\Counterfactual Explanations
	in Recommender Systems}

\author{Azin Ghazimatin}
\affiliation{%
	\institution{Max Planck Institute for Informatics, Germany}
}
\email{aghazima@mpi-inf.mpg.de}

\author{Oana Balalau}
\authornote{This work was done while the author was at the MPI for
	Informatics.}
\affiliation{%
	\institution{Inria and \'Ecole Polytechnique, France}
	}
\email{oana.balalau@inria.fr}

\author{Rishiraj Saha Roy}
\affiliation{%
	\institution{Max Planck Institute for Informatics, Germany}
}
\email{rishiraj@mpi-inf.mpg.de}

\author{Gerhard Weikum}
\affiliation{%
	\institution{Max Planck Institute for Informatics, Germany}
}
\email{weikum@mpi-inf.mpg.de}

{\tiny }\renewcommand{\shortauthors}{Ghazimatin et al.}

\newcommand\BibTeX{B{\sc ib}\TeX}

\newcommand{\squishlist}{
 \begin{list}{$\bullet$}
  { \setlength{\itemsep}{0pt}
     \setlength{\parsep}{3pt}
     \setlength{\topsep}{3pt}
     \setlength{\partopsep}{0pt}
     \setlength{\leftmargin}{1.5em}
     \setlength{\labelwidth}{1em}
     \setlength{\labelsep}{0.5em} } }

\newcommand{\squishend}{
  \end{list}  }

\begin{abstract}
Interpretable explanations for recommender systems and other machine learning
models are crucial to gain user trust. Prior works that have focused on paths
connecting users and items in a heterogeneous network have several limitations,
such as discovering relationships rather than true explanations, or disregarding
other users' privacy.
In this work, we take a fresh perspective, and present \prince: a provider-side
mechanism to produce tangible explanations for end-users, where an explanation is
defined to be
\textit{a set of minimal actions performed by the user}
that, if removed,
changes the recommendation to a different item.
Given a recommendation, \prince uses a polynomial-time optimal algorithm for
finding this minimal set of a user's actions from an exponential search space,
based on random walks over dynamic graphs. Experiments on two real-world datasets
show that \prince provides more compact explanations than intuitive baselines, and
insights from a crowdsourced user-study demonstrate the viability of such
action-based explanations. We thus posit that \prince produces \textit{scrutable},\
\textit{actionable},
and \textit{concise explanations}, owing to its use of
\textit{counterfactual} evidence,
a user's \textit{own actions},
and \textit{minimal} sets, respectively.
\end{abstract}

\ccsdesc[500]{Information systems~Recommender systems}


\maketitle

\section{Introduction}
\label{sec:intro}

\noindent\textbf{Motivation.} Providing user-comprehensible explanations for
machine learning models
has 
gained
prominence in multiple
communities~\cite{zhang2019ears,miller2019xai,rimchala2019kdd,zhang2019cvpr}.
Several 
studies have shown that explanations 
increase users' trust in systems that
generate personalized recommendations or other rankings
(in news, entertainment, etc.)
~\cite{ribeiro2016should,kunkel2019let,kouki2019personalized}.
%
Recommenders have become very sophisticated, exploiting signals from
a complex interplay of factors
like users' activities, interests and social links~\cite{zhang2018explainable}.
Hence the pressing need for explanations.

Explanations for recommenders
can take
several forms, 
depending on the generator (\textit{explanations by whom?}) and
the consumer (\textit{explanations for whom?}). As generators, only
\textit{service providers} can produce 
true explanations for how systems compute
the recommended items~\cite{balog2019transparent,zhang2014explicit,
	wang2018explainable};
\textit{third parties} can merely discover relationships
and create post-hoc rationalizations for black-box models that may look
convincing to 
users~\cite{ghazimatin:19,peake2018explanation,wang2018reinforcement}.
On the consumer side, \textit{end-users} can grasp tangible aspects like
activities, likes/dislikes/ratings or demographic factors.
Unlike {system developers} or accountability engineers, end-users would obtain 
hardly any insight from transparency of internal system workings.
In this work, we deal with explanations
\textit{by the provider and for the end-user}.

\begin{figure} [t]
	\centering
	\includegraphics[width=\columnwidth]{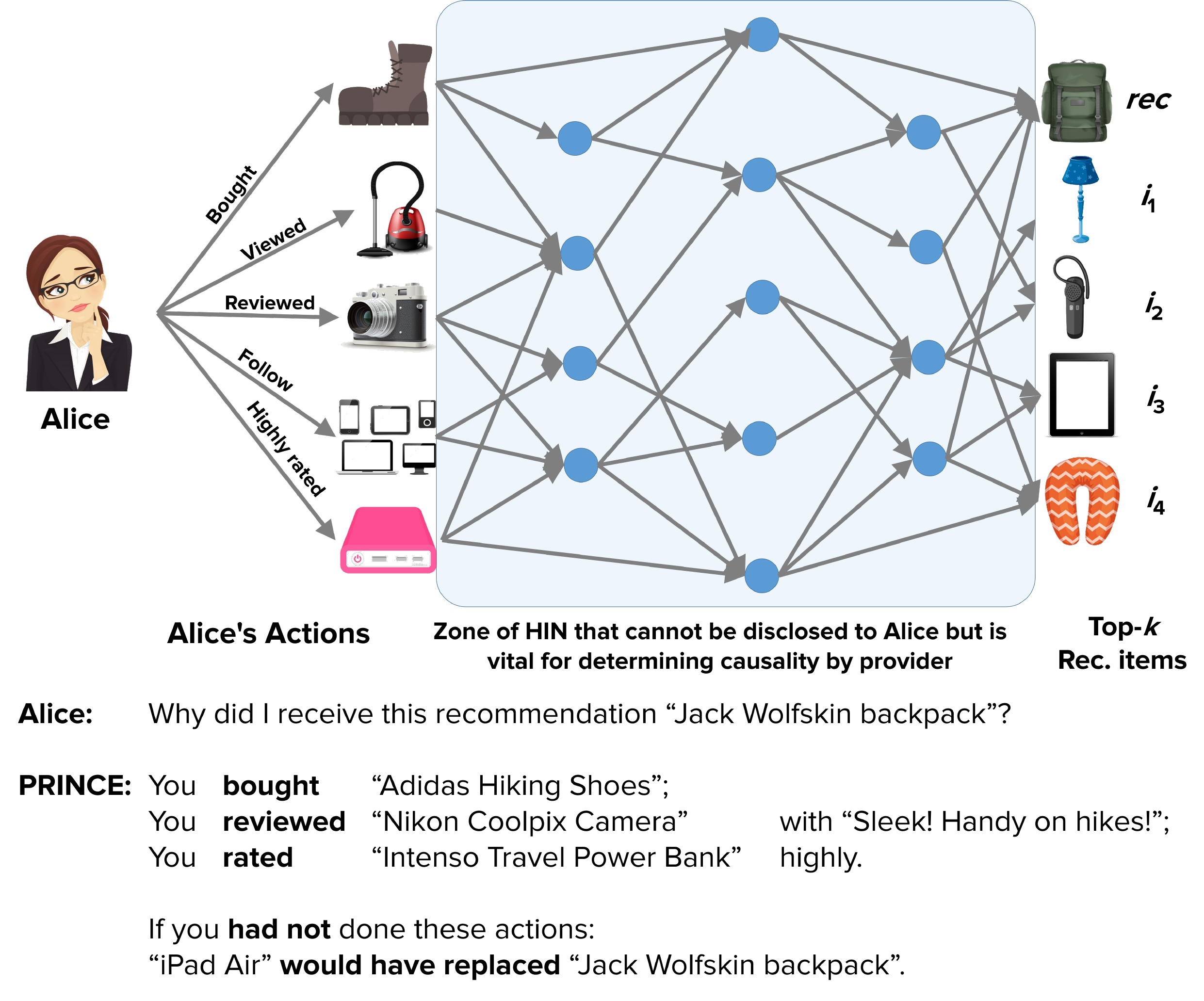}
	\caption{\prince generates explanations as a minimal set of actions
		using counterfactual evidence on user-specific HINs.}
	\label{fig:pixie}
	\vspace*{-0.7cm}
\end{figure}

\vspace{0.3cm}
\noindent\textbf{Limitations of state-of-the-art.} At the core of most recommender
systems 
is some variant of matrix or tensor decomposition (e.g., \cite{koren2009matrix})
or spectral graph analysis
(e.g., \cite{DBLP:conf/kdd/JamaliE09}), with various forms of regularization
and often involving gradient-descent methods for parameter learning.
One of the recent and popular paradigms is based on 
\textit{heterogeneous information networks (HIN)}
\cite{yu2014personalized,yu2013recommendation,
	DBLP:journals/tkde/ShiLZSY17,zhang2019shne},
a powerful model
that represents relevant entities and actions
as a directed and weighted graph with multiple
node and edge types. 
Prior efforts towards explanations for HIN-based recommendations have mostly
focused on \textit{paths} that connect the user with the recommended item
~\cite{shi2015semantic,ghazimatin:19,ai2018learning,wang2018ripplenet,
wang2019explainable,xian2019reinforcement,DBLP:conf/icdm/YangLWH18}.
An application of path-based explanations, for an online shop, would
be of the form: 
\squishlist
\item[ ] User $u$ received item $rec$ because $u$ follows user $v$,
who bought item $j$,
which has the same category as $rec$.
\squishend
However, such methods come with critical
privacy concerns arising from nodes in paths that disclose other users'
actions or interests to user $u$, like the purchase of user $v$
above.
%
Even if user $v$'s id was anonymized, user $u$ would know whom
she is following and could often guess who user $v$ actually is,
that bought item $j$,
assuming that $u$ has a relatively small set
of followees~\cite{machanavajjhala2011personalized}.
If entire paths containing other users are suppressed instead, then
such explanations would no longer be faithful to the true cause.
Another family of
path-based methods~\cite{ghazimatin:19,peake2018explanation,wang2018reinforcement}
presents plausible connections
between users and items as justifications. 
However, this is merely post-hoc rationalization,
and not actual causality.


\vspace{0.3cm}
\noindent\textbf{Approach.} This paper presents \prince, 
a method for 
\textbf{Pr}ovider-side \textbf{In}terpretability with \textbf{C}ounterfactual
 \textbf{E}vidence, that
overcomes 
the outlined limitations.
\prince is a provider-side solution
aimed at detecting the actual cause responsible for the recommendation, in 
a heterogeneous information network with users, items, reviews, and categories.
\prince's explanations are grounded in the user's own actions, and
thus preclude privacy concerns of path-based models.
 Fig.~\ref{fig:pixie} shows an illustrative example. Here, 
Alice's actions like bought shoes, reviewed a camera, and rated
a power bank are deemed as explanations for her
backpack recommendation.
One way of identifying a user's actions for an explanation would be
to compute scores of actions with regard to the recommended item.
However, this would be an unwieldy distribution over potentially
hundreds of actions -- hardly comprehensible to an end-user. 
Instead, we operate in
a \textit{counterfactual} setup~\cite{martens2014explaining}.
\prince
identifies a small (and actually minimal) set
of a user's actions such that removing these actions would result in
replacing the recommended item with a different item.
In Fig.~\ref{fig:pixie}, the item \textit{rec =} ``Jack Wolfskin backpack''
would be replaced, as the system's top recommendation,
by $i_3=$``iPad Air'' (the $i$'s represent candidate replacement items).
Note that there may be multiple such minimal sets, but uniqueness is not
a concern here. 

Another perspective here is that the goal of an explanation is often
to show users
\textit{what they can do}
in order to receive more relevant recommendations. Under this claim, the end-user
has no \textit{control} on the network beyond her immediate neighborhood, i.e.,
the network beyond is \textit{not actionable} (shaded zone in
Fig.~\ref{fig:pixie}), motivating \prince's choice
of grounding explanations in users' own actions.

For true explanations, we need to commit ourselves to a specific family
of recommender models. In this work, we choose a general framework
based on \textit{Personalized PageRank
(PPR)}, as used in the state-of-the-art RecWalk
system~\cite{nikolakopoulos2019recwalk}, and adapt it to
the HIN setup. 
The heart of \prince is a polynomial-time algorithm for exploring
the (potentially exponential) search space of subsets of user actions --
the candidates for
causing the recommendation.
The algorithm efficiently computes PPR contributions
for groups of actions with regard to an item, by adapting the
reverse
local push algorithm of \cite{andersen2007local} 
to a dynamic graph setting~\cite{zhang2016approximate}.
In summary, the desiderata for the explanations from \prince (in \textbf{bold})
connect to the technical approaches adopted (in \textit{italics})
in the following ways. Our explanations are:
\squishlist
\item \textbf{Scrutable}, as they are derived in a \textit{counterfactual} setup;
\item \textbf{Actionable}, as they are grounded in the user's
own \textit{actions};
\item \textbf{Concise}, as they are \textit{minimal} sets changing
a recommendation.
\squishend

Extensive experiments with Amazon and Goodreads datasets show 
that \prince{}'s minimal explanations, achieving the desired
item-replacement effect, cannot be easily obtained by heuristic methods based
on contribution scores and shortest paths.
A crowdsourced user study on Amazon Mechanical Turk (AMT) provides additional
evidence that \prince{}'s explanations 
are more useful 
than ones based on paths~\cite{DBLP:conf/icdm/YangLWH18}.
Our code is public at \url{https://github.com/azinmatin/prince/}.

\vspace{0.3cm}
\noindent\textbf{Contributions.} Our salient contributions in this work are:
\squishlist
	\item \prince is the first work that explores counterfactual evidence
	for discovering causal explanations in a heterogeneous information network;
	\item \prince is the first work that defines explanations for recommenders
	in terms of users' own actions; 
	\item We present an optimal algorithm that explores the search space
	of action subsets in polynomial time, for efficient computation
of a minimal subset of user actions;
	\item Experiments with two large datasets and a user study
show that \prince  can effectively aid a service provider
in generating user-comprehensible causal explanations
for recommended items.
\squishend

\vspace*{-0.2cm}
\section{Computational Model}
\label{sec:model}

\noindent{\bf Heterogeneous Information Networks (HIN).}
A \emph{heterogeneous graph} $G = (V, E, \theta)$ consists of a set of
nodes $V$, a set of edges $E \subseteq V \times V,$ and a mapping $\theta$
from each node and each edge to their types, such that
$\theta_V: V \rightarrow T_V$ and $\theta_E: E \rightarrow T_E$ with
$|T_V| + |T_E| > 2$. In our work, a heterogenous graph contains at least
two node types, users $U \in T_V$ and items $I \in T_V$. For simplicity,
we use the notations $U$ and $I$ to refer both to the type of a node and
the set of all nodes of that type. A  graph is \emph{weighted} if there is
a weight assigned to each edge, $w: E \rightarrow \mathbb{R}$, and a graph is
\emph{directed} if $E $ is a set of ordered pairs of nodes. We denote with
$N_{out}(v)$ and  $N_{in}(v)$ the sets of out-neighbors and  
in-neighbors of node $v$, respectively.  A directed and weighted heterogeneous
graph where 
each node $v \in V$ and each edge $e \in E$  belong to exactly one type,
is  called a \emph{heterogenous information network
	(HIN)}~\cite{DBLP:journals/tkde/ShiLZSY17}.

\vspace{0.2cm}
\noindent{\bf Personalized PageRank (PPR) for recommenders.}
We use Personalized PageRank (PPR) 
for recommendation in
HINs~\cite{haveliwala2003topic,nikolakopoulos2019recwalk}.
PPR is the stationary distribution of a random walk in $G$ in which,
at a given step, with probability $\alpha$, a surfer teleports to a 
set of seed nodes $\{s\}$, and with probability $1 - \alpha$, continues
the walk to a randomly chosen outgoing edge from the current node. 
More precisely, given $G$, teleportation probability $\alpha$,
a single seed $s$, the one-hot vector $e_s$, and the 
transition matrix $W$, the Personalized PageRank vector $PPR(s)$ is defined
recursively as:
\begin{equation}
	PPR(s, \cdot) = \alpha e_s + (1-\alpha)PPR(s, \cdot)W
	\label{eq:ppr}
\end{equation}
Let $PPR(s,v)$ be the PPR score of node $v$ personalized for $s$.
We define the \emph{PPR recommendation} for user $u \in U$, or
the top-$1$ recommendation, as:
\begin{equation}
	rec = \argmax_{i \in I \setminus N_{out}(u)} PPR(u, i)
	\label{eq:rec}
\end{equation}
Given a set of edges $A \subset E$, we use the notation $PPR(u, i|A)$ to define
the PPR 
of an item $i$ personalized for a user $u$ in
the graph $G = (V, E \setminus A, \theta)$. 
We refer to this graph as $G \setminus A$.
To improve top-$n$ recommendations,
Nikolakopoulos et al.~\cite{nikolakopoulos2019recwalk} define a random walk
in an HIN $G$ as follows: 
\begin{itemize}
\item With probability $\alpha$, the surfer teleports to $u$
\item With probability $1- \alpha$, the surfer continues the walk in
the following manner:
	\begin{itemize}
	\item[+] With probability $1 - \beta$, the random surfer 
	moves
	to a \textit{node of the same type}, using a 
	similarity-based stochastic 
	transition matrix
	\item[+] With probability $\beta$, the surfer chooses
	any outgoing edge at random.
	\end{itemize}
\end{itemize}

For each node type $t$ in $T_V$, there is an associated stochastic similarity
matrix $S_t$, which encodes the relationship between the nodes of type $t$.
When nodes of the same type are not comparable, the similarity matrix is
the identity matrix, i.e. $S_t = I$. 
Otherwise, an entry $(i,j)$ in $S_t$ corresponds to the similarity between
node $i$ and
node $j$. The stochastic process described by this walk is
a nearly uncoupled Markov chain~\cite{nikolakopoulos2019recwalk}.
The stationary distribution of the random walk is the PPR with
teleportation probability $\alpha$ in a graph $G^{\beta}$
(referred to as $RecWalk$ in~\cite{nikolakopoulos2019recwalk}), where
the  transition probability matrix of $G^{\beta}$ is:
\begin{equation}
	W^{\beta} =  \beta W + (1 - \beta) S
	\label{eq:W}
\end{equation}
The matrix $W$ is the transition probability matrix of
the original graph $G$. Matrix $S = Diag(S_{1}, S_{2}, \cdots, S_{|T_V|})$
is a diagonal matrix of
order $|V|$. 

\vspace{0.2cm}
\noindent{\bf Counterfactual Explanations.}
A user $u$ interacts with items via different types of \emph{actions} $A$,
such as 
clicks, purchases, ratings or reviews, which  are captured as
\emph{interaction edges} in the graph $G$. 
Our goal is to present user $u$ with a set of interaction edges $A^* 
\subseteq
\{(u, n_i)| (u, n_i) \in A\}$
(where $n_i$ is a neighbor of $u$)
responsible for an item recommendation $rec$;
we refer to this as a \emph{counterfactual explanation}. 
An explanation is \emph{counterfactual}, if after removing the edges $A^*$
from the graph, the user receives a different top-ranked recommendation
$rec^*$. 
A counterfactual explanation $A^*$ is \emph{minimal} 
if there is no smaller set $A' \subseteq A$
such that
$|A'| < |A^*|$ and
$A'$ is also a counterfactual explanation for $rec$.

\vspace{0.2cm}
\noindent \textbf{Formal problem statement.} Given 
a heterogenous information network $G = (V, E, \theta)$ and 
the top-ranked recommendation $rec \in I$ for user $u \in U$,
find a minimum counterfactual explanation for $rec$.

\section{The \prince Algorithm}
\label{subsec:algo}

\begin{figure*}[t]
	\begin{subfigure}[t]{0.3\textwidth}
		\includegraphics[width=\textwidth]{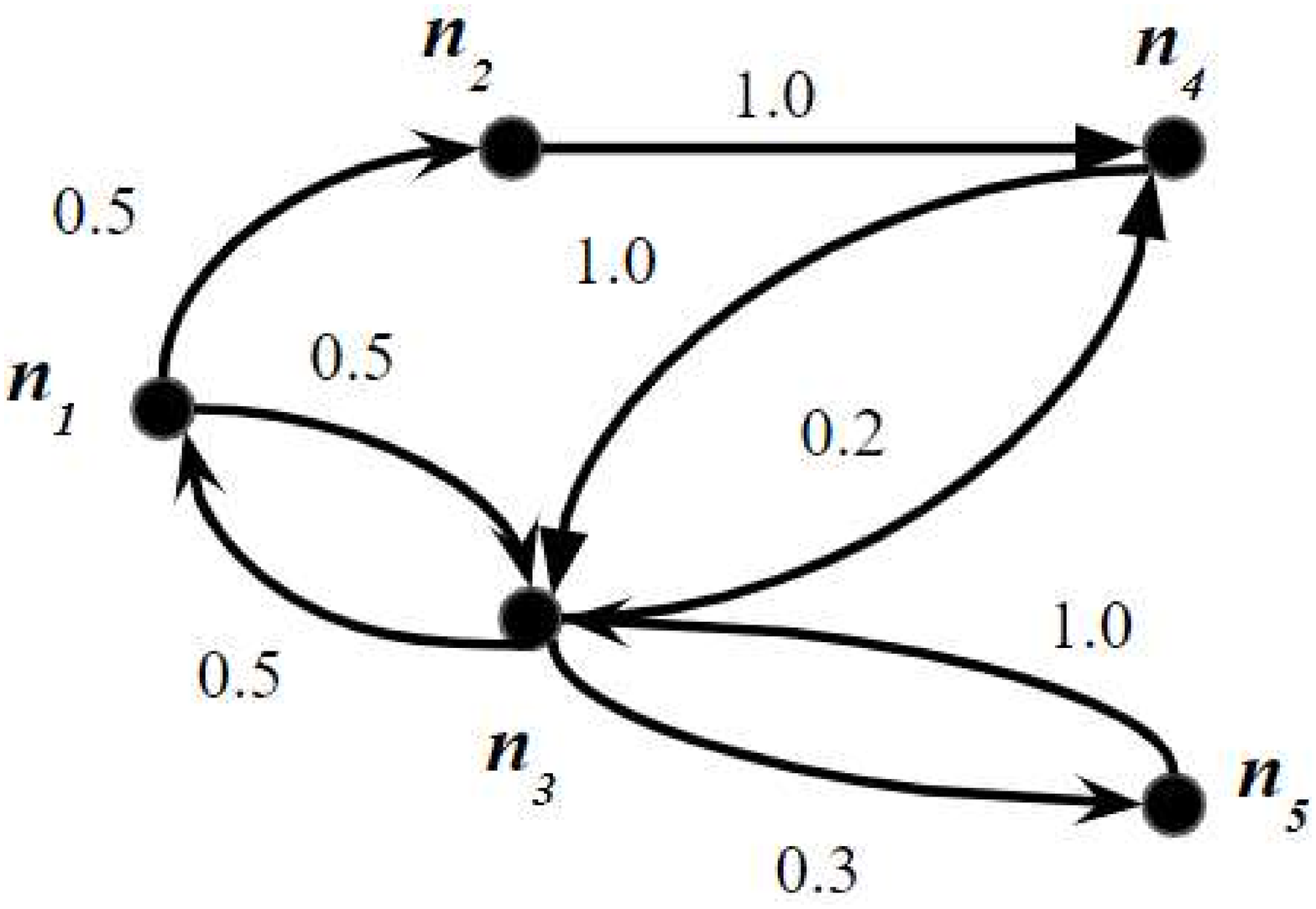}
		\caption{\scriptsize{$PPR(n_1, n_4) = 0.160
				\\ PPR(n_1, n_5) = 0.085
				\\ PPR(n_1, n_4) > PPR(n_1, n_5)$}}
		\label{fig-a}
	\end{subfigure}\hfill
	\begin{subfigure}[t]{0.3\textwidth}
		\includegraphics[width=\textwidth]{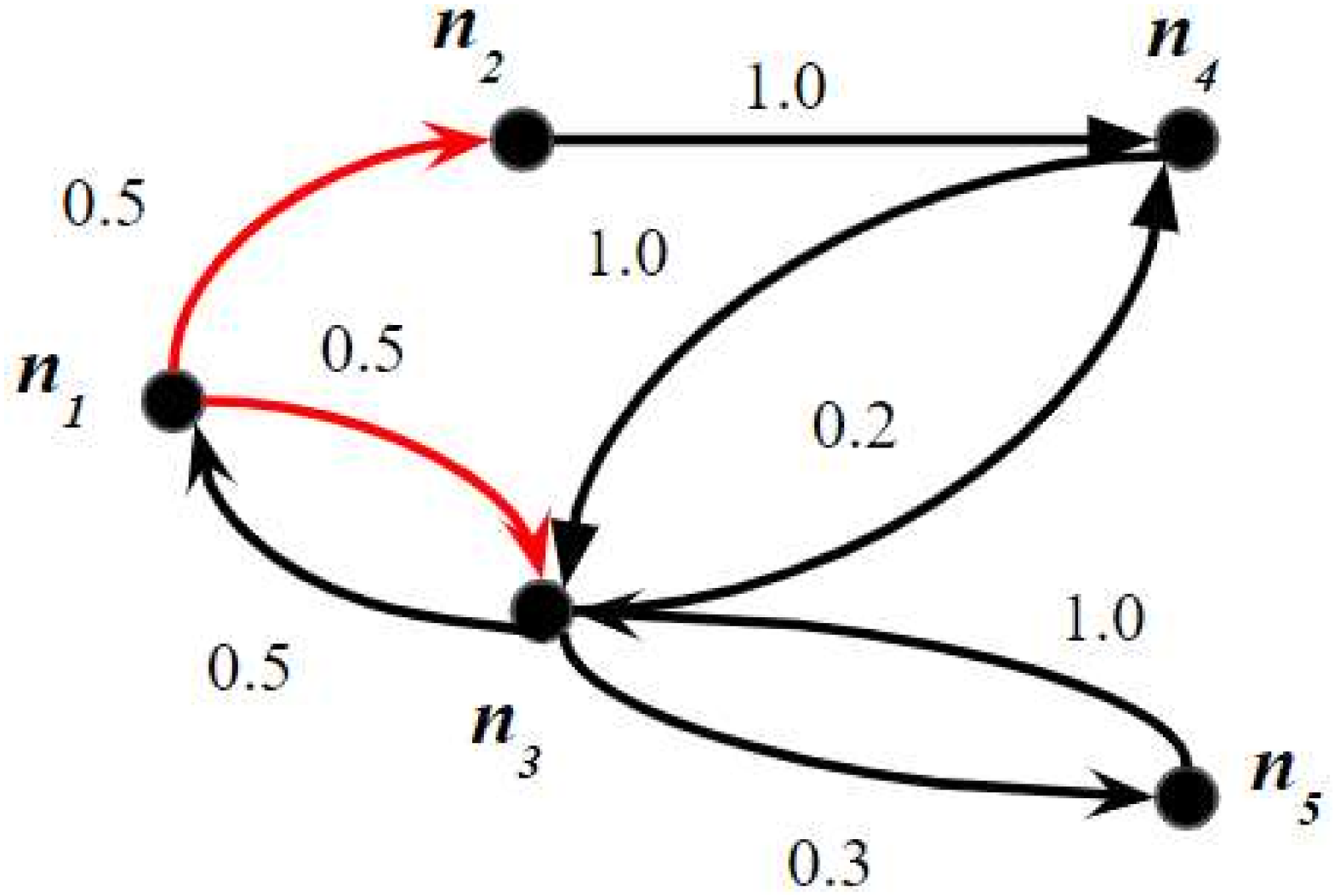}
		\caption{\scriptsize{$A = \{(n_1, n_2), (n_1, n_3)\}
				\\ W(n_1 , n_2) [PPR(n_2, n_4| A) - PPR(n_2, n_5 | A)] = 0.095
				\\ W(n_1 , n_3) [PPR(n_3, n_4 | A) - PPR(n_3, n_5 | A)] = -0.022$}}
		\label{fig-b}
	\end{subfigure}\hfill
	\begin{subfigure}[t]{0.3\textwidth}
		\includegraphics[width=\textwidth]{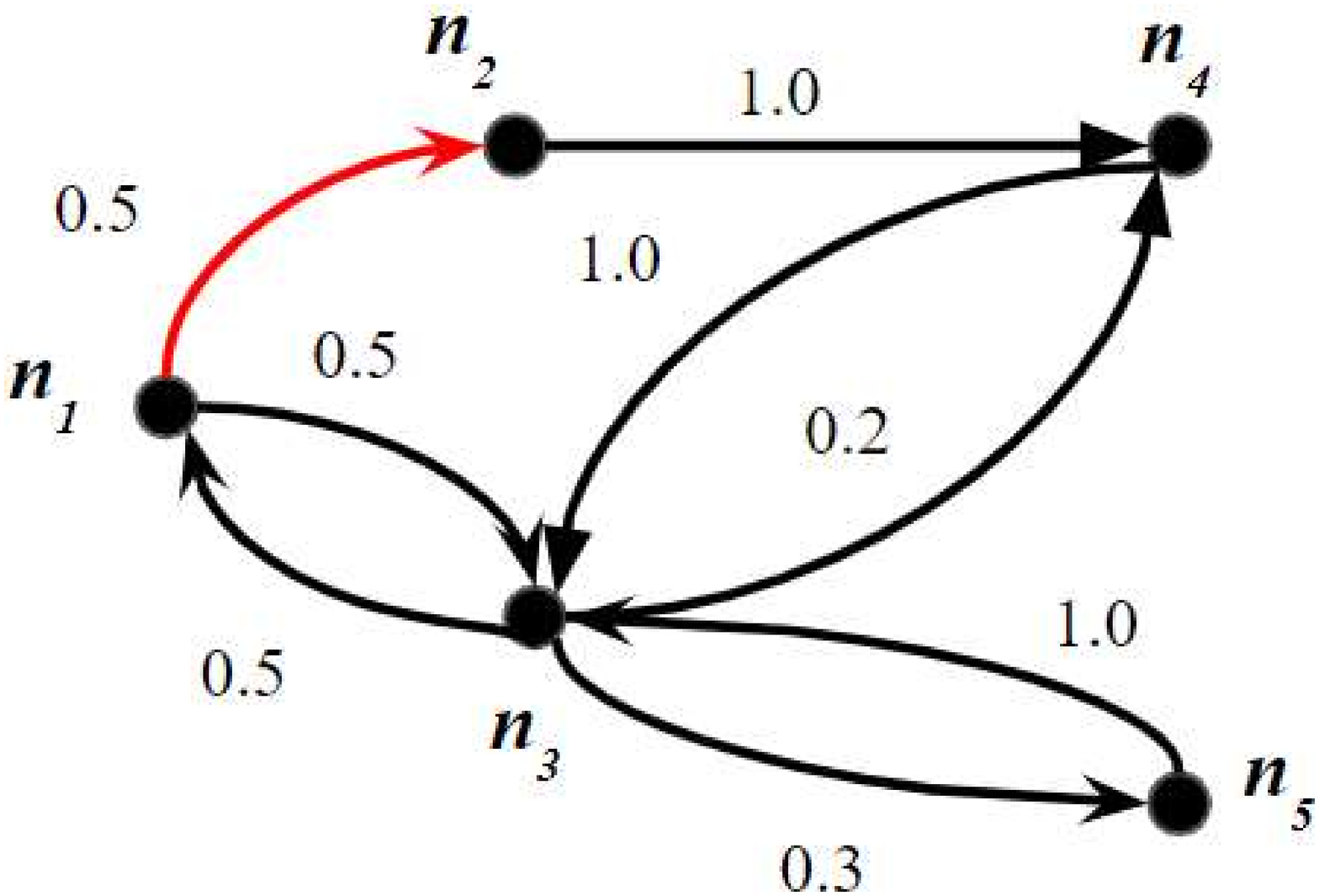}
		\caption{\scriptsize{$ A^*  = \{(n_1, n_2)\}
				\\ PPR(n_1, n_4 | A^*) = 0.078
				\\ PPR(n_1, n_5 | A^*) = 0.110
				\\ PPR(n_1, n_5 | A^*) > PPR(n_1 , n_4| A^*)$}
		}
		\label{fig-b}
	\end{subfigure}
	\caption{Toy Example. (a) A weighted and directed graph where
		the $PPR$ scores are personalized for node $n_1$. Node $n_4$ has
		higher $PPR$ than $n_5$.
		(b) Scores in a graph configuration where outgoing edges
		$(n_1, n_2)$, and $(n_1, n_3)$ are removed (marked in red).
		(c) Removing $(n_1, n_2)$ causes $n_5$ to outrank $n_4$.} 
	\vspace*{-0.3cm}
	\label{fig:pixie-method}
\end{figure*}

\setlength{\textfloatsep}{0.1cm}
\setlength{\floatsep}{0.1cm}
\begin{algorithm} [t]
	\small
	\DontPrintSemicolon
	\SetAlgoLined
	\SetKwInput{Input}{Input}\SetKwInput{Output}{Output}
	\SetKwComment{Comment}{$\triangleright$\ }{}
	\Input{$G=(V, E, \theta)$, $I \subset V$, $u \in V$, $rec \in I$}
	\Output{$A^*$ for $(u, rec)$}
	$A^* \gets A$ \;
	$rec^* \gets rec$\;
	
	\ForEach {$i \in I$} {\label{alg:mainstart}
		$A^i\gets \texttt{SwapOrder}(G, u, rec, i)$ \;
		\tcp*[l]{Actions $A^i$ swap orders of $rec$ and $i$}
		\If { $|A^i| < |A^*|$}
		{
			$A^* \gets A^i$ \;
			$rec^* \gets i$
		}
	
	\ElseIf {$|A^i| = |A^*|$ and $PPR(u, i|A^i) > PPR(u, rec^*|A^i)$}
	{
		\label{alg:repstart}
		$A^* \gets A^i$ \;
		$rec^* \gets i$
		\label{alg:repend}
	}
	}\label{alg:mainend}

	\Return $A^*, rec^*$ \;
	\BlankLine
	\SetKwFunction{FMain}{SwapOrder}
	\SetKwProg{Fn}{Function}{:}{end}
	\Fn{\FMain{$G, u, rec, rec^*$}}{
		$A \gets \{(u, n_i)| n_i \in N_{out}(u), n_i \neq u\}$ \;
		$A^* \gets \emptyset$ \;
		$H\gets \text{MaxHeap}(\phi)$ \;
		$sum \gets 0 $ \;
		\ForEach {$(u, n_i) \in A$} { \label{alg:sortstart}
			$diff \gets W(u, n_i) \cdot (PPR(n_i, rec|A) - PPR(n_i, rec^*|A))$\;
			$H.insert(n_i, diff)$ \;
			$sum \gets sum + diff$
		}\label{alg:sortend}
		\While {$sum > 0 \text{ and } |H| > 0$} {\label{alg:updatestart}
			$(n_i, diff) \gets H.delete\_max()$ \;
			$sum \gets sum - diff$ \;
			$ A^* \gets  A^* \cup (u, n_i)$
		}\label{alg:updateend}
		\lIf {$sum > 0$} {$A^* \gets A$}
		\Return $A^*$
	}
	\caption{\prince}
\label{alg:poly}	
\end{algorithm}	
\setlength{\textfloatsep}{0.1cm}
\setlength{\floatsep}{0.1cm}
	
In this section, we develop an algorithm for
computing a minimum counterfactual explanation for user $u$
receiving recommended item $rec$, given the PPR-based
recommender framework $RecWalk$~\cite{nikolakopoulos2019recwalk}. 
A na\"{i}ve optimal algorithm enumerates all subsets of
actions $A^* \subseteq A$, 
and checks whether the removal of each of these subsets
replaces $rec$ with
a different item as the top recommendation,
and finally selects the subset with the minimum size.
This approach is exponential in the number of actions of the user. 

To devise a more efficient and practically viable algorithm,
we express the $PPR$ scores as follows~\cite{jeh2003scaling}, with
$PPR(u,rec)$ denoting the PPR of $rec$
personalized for $u$ (i.e., jumping back to $u$):
\begin{equation}
	PPR(u, rec) = (1-\alpha)\sum_{n_i \in N_{out}(u)} W(u, n_i)
	PPR(n_i, rec) + \alpha \delta_{u,rec}   
	\label{eq:invariant}
\end{equation}
where $\alpha$ denotes the teleportation probability (probability 
of jumping back to $u$) and $\delta$ is the Kronecker delta function. The only
required modification, with regard to 
$RecWalk$~\cite{nikolakopoulos2019recwalk}, is the transformation of
the transition probability matrix from $W$ to $W^{\beta}$. 
For simplicity, we will refer to the adjusted probability matrix as $W$.

Eq.~\ref{eq:invariant} shows that the PPR of $rec$ personalized for user $u$,
$PPR(u,rec)$, is a function of the PPR values of $rec$ personalized for
the neighbors of $u$. 
Hence, in order to decrease $PPR(u, rec)$, 
we can remove edges $(u, n_i), n_i \in N_{out}(u)$.
To replace the recommendation $rec$ with 
a different item $rec^*$, a simple heuristic 
would remove edges $(u, n_i)$ in non-increasing order of their contributions
$W(u, n_i) \cdot PPR(n_i, rec)$. 
However, although this would reduce the PPR of $rec$,
it also affects and possibly reduces the PPR of other items, too,
due to the recursive nature of PPR, where all paths matter.
%

 
Let $A$ be the set of outgoing edges of a user $u$ and let $A^*$
be a subset of $A$, such that $A^* \subseteq A$. The main intuition behind our 
algorithm is that we can express
$PPR(u, rec)$ after the removal of $A^*$,
denoted by $PPR(u, rec|A^*)$, as a function of two 
components: $PPR(u, u|A^*)$ and the values $ PPR(n_i, rec|A)$,
where $n_i \in \{ n_i| (u, n_i) \in A \setminus A^*\}$
and $n_i \neq u$. 
The score $PPR(u, u|A^*)$ does not depend on $rec$, and
the score $PPR(n_i, rec|A)$ is independent of
$A^*$. 

Based on these considerations, we present Algorithm~\ref{alg:poly}, 
proving its correctness in Sec.~\ref{sec:correct}.
Algorithm~\ref{alg:poly} takes as input a graph $G$, a user $u$,
a recommendation $rec$, and a set of items $I$.
In lines~\ref{alg:mainstart}-\ref{alg:mainend}, we iterate through
the items $I$, and find the minimum counterfactual explanation $A^*$.
Here, $A^i$ refers to 
the actions whose removal swaps the orders of items $rec$ and $i$.
In addition, we ensure that after removing $A^*$, we return the item with
the highest PPR score as the replacement item
(lines~\ref{alg:repstart}-\ref{alg:repend}).
Note that in the next section, we propose an equivalent formulation for
the condition $PPR(u, i | A^i) > PPR(u, rec^* | A^i)$, eliminating the need
for recomputing scores in $G \setminus A^*$.   

The core of our algorithm is
the function \texttt{SwapOrder}, which receives as input two items,
$rec$ and $rec^*$, and a user $u$. In
lines \ref{alg:sortstart}-\ref{alg:sortend}, we sort
the interaction edges $(u, n_i) \in A$ in non-increasing
order of their contributions
$W(u, n_i) \cdot (PPR(n_i, rec|A) - PPR(n_i, rec^*|A))$.
In lines \ref{alg:updatestart}-\ref{alg:updateend}, we remove
at each step, the outgoing interaction edge with the highest contribution,
and update $sum$ and $A^*$ correspondingly. The variable $sum$ is strictly positive
if in the current graph configuration ($G \setminus A^*$),
$ PPR(u, rec) > PPR(u, rec^*)$.
This 
constitutes the main building block of our approach.
Fig.~\ref{fig:pixie-method} illustrates the execution of
Algorithm~\ref{alg:poly} on a toy example. 		

The time complexity of the algorithm is $O(|I| \times |A| \times \log |A|)$,
plus the cost of computing PPR for these nodes.
The key to avoiding the exponential cost of considering
all subsets of $A$ is the insight that \textit{we need only to compute
PPR values for alternative items with personalization 
based on a graph where all user actions $A$ are removed}.
This is feasible because the action deletions affect only
outgoing edges of the teleportation target $u$,
as elaborated 
in Sec.~\ref{sec:correct}.

The PPR computation could simply re-run a power-iteration
algorithm for the entire graph, or compute the principal
eigenvector for the underlying matrix. 
This could be cubic in the graph size (e.g., if we use
full-fledged SVD), but it keeps us in the regime of
polynomial runtimes. 
In our experiments, we use the much more efficient
reverse local push algorithm~\cite{andersen2007local} 
for PPR calculations.


%


\section{Correctness Proof}
\label{sec:correct}

We prove two main results:
\squishlist
\item[(i)] $PPR(u, rec|A^*)$  can be computed as a product of two
components where one depends on the modified graph
with the edge set $E \setminus A$ (i.e., removing all user
actions) and the other depends on the choice of
$A^*$ but not on the choice of $rec$.
\item[(ii)] To determine if some $A^*$ replaces
the top node $rec$ with a different node $rec^*$
which is not an out-neighbor of $u$, we need to
compute only the first of the two components in (i).
\squishend
\begin{theorem}
	Given a graph $G = (V, E)$, a node $u$ with outgoing edges
	$A$ such that $(u, u) \notin A$, a set of
	edges $A^* \subset A$, a node $rec \notin N_{out}(u)$, the PPR of $rec$
	personalized for $u$ in 
	the modified graph $G^* = (V, E \setminus A^*)$ can be expressed as follows:
	$$ PPR(u, rec|A^*) = PPR(u, u|A^*) \cdot f\big(\big\{PPR(n_i, rec|A)\big|
	(u, n_i) \in A \setminus A^*\big\}\big) $$
	where $f(\cdot)$ is an aggregation function. 
	\label{thm: pprs-rel}
\end{theorem}
\vspace*{-0.3cm}
\begin{proof}
Assuming that each node has at least one outgoing edge, the PPR can 
be expressed as the sum over the probabilities of 
walks of length $l$ starting at a node $u$~\cite{andersen2006local}:
	\begin{equation}
	PPR(u, \cdot) = \alpha\sum_{l=0}^\infty (1-\alpha)^l e_{u} W^l 
	\label{eq: ppr-walks}
	\end{equation}
	where $e_{u}$ is the one-hot vector for $u$. 
To analyze the effect of deleting $A^*$, we split the walks from $u$ to
	$rec$ into two parts, ($i$) the part representing the sum over
	probabilities of walks that start at $u$ and pass again by $u$,
	which is equivalent to $\alpha^{-1}PPR(u, u|A^*)$ (division by $\alpha$
	is required as the walk does not stop at $u$), and ($ii$) the part
	representing the sum over probabilities of walks starting at node $u$
	and ending at $rec$ without revisiting $u$ again,
	denoted by $p_{-u}(u, rec | A^*)$. 
	Combining these constituent parts, PPR can be stated as follows:
	\begin{equation}
	PPR(u, rec | A^*) = \alpha^{-1} PPR(u, u | A^*) \cdot p_{-u}(u, rec | A^*)
	\label{eq: new-ppr-0}
	\end{equation}
	

As stated previously, $p_{-u}(u, rec | A^*)$ represents the sum over
the probabilities of the walks from $u$ to $rec$ without revisiting $u$.
We can
express these walks using the remaining neighbors of $u$
after removing $A^*$:
\begin{equation}
	p_{-u}(u, rec | A^*) = (1 - \alpha)
	\sum_{(u, n_i) \in A \setminus A^*} W(u, n_i) \cdot p_{-u}(n_i, rec | A^*)
\end{equation}
where $p_{-u}(n_i, rec | A^*)$ refers to the walks starting at $n_i$ ($n_i \neq u$) 
and ending at $rec$ that do not visit $u$.
We replace $p_{-u}(n_i, rec | A^*)$ with its equivalent formulation 
$PPR(n_i, rec | A)$. $PPR(n_i, rec)$ in graph $G\setminus A$ is computed as
the sum over the probabilities of walks that never pass by $u$.
Eq.~\ref{eq: new-ppr-0} can be rewritten as follows:	 
\begin{align}
	& PPR(u, rec|A^*) \notag \\
	& =  PPR(u, u|A^*) \cdot \alpha^{-1} (1-\alpha)
	\sum_{(u, n_i) \in A \setminus A^*}
	W(u, n_i|A^*) PPR(n_i, rec|A)
\end{align}
This equation directly implies:
\begin{equation}
	PPR(u, rec|A^*) = PPR(u, u|A^*) \cdot f\big(\big\{PPR(n_i, rec|A)\big|
	(u, n_i) \in A \setminus A^*\big\}\big)
\end{equation}
\end{proof}
\vspace*{-0.3cm}
\begin{theorem}
The minimum counterfactual explanation for $(u, rec)$ 
can be computed in polynomial time. 
\label{thm: poly}
\end{theorem}
\vspace*{-0.3cm}
\begin{proof}
	

We show that there exists
a polynomial-time algorithm for finding the minimum set $A^* \subset A$
such that $ PPR(u, rec|A^*) < PPR (u, rec^*|A^*)$, if such a set exists. 
Using Theorem~\ref{thm: pprs-rel}, we show that one can compute if some
$rec^*$ can replace the original $rec$ as the top
recommendation, solely based on PPR scores from a single graph
where all user actions $A$ are removed:
\begin{align}
	 & PPR(u, rec|A^*) < PPR (u, rec^*|A^*) \notag \\
	 & \Leftrightarrow \sum_{(u, n_i) \in A \setminus A^*}  W(u, n_i|A^*)
	 \big(PPR(n_i, rec|A)  -  PPR(n_i, rec^*|A) \big) < 0  \notag \\
	 & \Leftrightarrow \sum_{(u, n_i) \in A \setminus A^*} W(u, n_i)\big
	 (PPR(n_i, rec|A)  -  PPR(n_i, rec^*|A) \big) < 0
	 \label{eq:equivalence}
\end{align}
	
The last equivalence is derived from:
\begin{equation}
	W(u, n_i|A^*) = \frac{W(u, n_i)}{1 - \sum_{(u, n_j) \in A^*} W(u, n_j)}
\end{equation}
	
For a fixed choice of $rec^*$, the summands in
expression~\ref{eq:equivalence}
do not depend on $A^*$, and so they are constants for all possible
choices of $A^*$.
Therefore, by sorting the summands in descending order, we can greedily
expand $A^*$ from a single action to many actions until some
$rec^*$ outranks $rec$. This approach is then guaranteed to arrive at
a minimum subset.

\end{proof}

\section{Graph Experiments}
\label{sec:graph-exp}

We now describe experiments performed with graph-based recommenders built from
real datasets to evaluate \prince.

\subsection{Setup}
\label{subsec:graph-setup}

\textbf{Datasets.} We used two 
real datasets: 
\squishlist
\item[(i)] The Amazon Customer Review
dataset (released by Amazon:
\url{s3.amazonaws.com/amazon-reviews-pds/readme.html}),
and,
\item[(ii)] The Goodreads review
dataset 
(crawled by the authors of~\cite{wan2018item}:
\url{sites.google.com/eng.ucsd.edu/ucsdbookgraph/home}).
\squishend

Each record in both datasets consists of a user, an item, its categories,
a review, and a rating value (on a $1-5$ scale).
In addition, a Goodreads data record has the book author(s) and
the book description.
We augmented the Goodreads collection with social links
(users following users) that we crawled from 
the Goodreads website.

The high diversity of categories in the Amazon data,
ranging from household equipment to food and toys, allows scope to
examine the interplay of cross-category information
within explanations.
The key reason for additionally choosing Goodreads is to 
include the effect of social connections (absent in the Amazon data). 
The datasets were converted to graphs with
``users'', ``items'', ``categories'', and ``reviews'' as \textit{nodes},
and ``rated'' (user-item),
``reviewed'' (user-item), ``has-review'' (item-review), 
``belongs-to'' (item-category) and
``follows'' (user-user) as \textit{edges}. 
In Goodreads, there is an additional node type ``author'' 
and an edge type ``has-author'' (item-author). All the edges, except 
the ones with type ``follows'', are bidirectional.  
Only ratings with value \textit{higher than three}
were considered, as low-rated items should not influence
further recommendations.

\begin{table} [t] 
	\centering
	\resizebox*{\columnwidth}{!}{
	\begin{tabular}{l c c c c c}
		\toprule
		\textbf{Dataset}	& \textbf{\#Users}	& \textbf{\#Items}	& \textbf{\#Reviews}	& \textbf{\#Categories}	& \textbf{\#Actions}	\\	\toprule
		Amazon				& $2k$				& $54k$				& $58k$					& $43$					& $114k$				\\ 
		Goodreads			& $1k$				& $17k$				& $20k$					& $16$					& $45k$					\\ \bottomrule
	\end{tabular}}
	\caption{Properties of the Amazon and Goodreads samples.}
	\label{tab:data}
	\vspace*{-0.3cm}
\end{table}
	
\textbf{Sampling.} For our experiments, we sampled $500$ seed users who had
between $10$ and $100$ actions, from both Amazon and Goodreads datasets.
The filters served to prune out
under-active and power users (potentially bots). 
Activity graphs were constructed
for the sampled users by taking their four-hop neighborhood from
the sampled data (Table~\ref{tab:data}).
Four is a reasonably small radius to keep the items relevant and personalized
to the seed users. 
On average, this resulted in having about $29k$ items
and $16k$ items for each user in their HIN, for Amazon
and Goodreads, respectively.

The graphs were augmented with weighted edges for
node similarity.
For Amazon, we added review-review edges where weights were computed
using the cosine similarity of the review 
embeddings, generated with Google's
Universal Sentence Encoder~\cite{cer2018universal},
with a cut-off threshold
$\tau = 0.85$ to retain only confident pairs.
This resulted in $194$ review-review edges. 
For Goodreads, we added three types of similarity edges: 
category-category, book-book and review-review, with the same
similarity measure 
($24$ category-category, $113$ book-book, and $1003$ 
review-review edges). Corresponding thresholds were
$0.67, 0.85$ and $0.95$.
We crawled category descriptions 
from the Goodreads' website and used book descriptions and review texts
from the raw data. Table~\ref{tab:data} gives some statistics about the
sampled datasets.

\textbf{Initialization.}
The replacement item for $rec$ is always chosen from the original top-$k$
recommendations generated by the system;
we systematically investigate the effect of $k$ on
the size of explanations in our experiments (with a default $k = 5$).
\prince 
does not need to be restricted to an explicitly specified candidate set,
and can actually operate over the full space of items $I$.
In practice, however, replacement items need to be guided by some 
measure of relevance to the user, or item-item similarity, so as not
to produce degenerate or trivial explanations if $rec$ is replaced
by some arbitrary item from a pool of thousands.

We use the standard teleportation probability
$\alpha = 0.15$~\cite{brin1998anatomy}. 
The parameter $\beta$ is set to $0.5$. 
To compute PPR scores, we used
the reverse local push method~\cite{zhang2016approximate} with 
$\epsilon=1.7e-08$ for Amazon and $\epsilon=2.7e-08$
for Goodreads. With these settings, \prince and the baselines 
were executed on all $500$ user-specific HINs to compute
an alternative recommendation (i.e., replacement item) $rec$*
and a counterfactual explanation set $A$*.	

\textbf{Baselines.}
Since
\prince
is an optimal algorithm with correctness
guarantees, it always finds minimal sets of actions that replace $rec$
(if they exist). We wanted to investigate, to what extent other, 
more heuristic, methods approximate the same effects.
To this end,
we compared \prince against two natural baselines:
\squishlist
\item[(i)] \textit{Highest Contributions} ($HC$): 
This is analogous to counterfactual evidence
in feature-based classifiers for structured
data~\cite{chen2017enhancing,moeyersoms2016explaining}.
It defines the contribution score of a user action $(u, n_i)$ to
the recommendation
score $PPR(u, rec)$ as $PPR(n_i, rec)$ (Eq.~\ref{eq:invariant}), and 
iteratively deletes edges with highest contributions until
the highest-ranked $rec$
changes to a different item.
\item[(ii)] \textit{Shortest Paths} ($SP$):
$SP$ computes the shortest path from $u$ to $rec$ and deletes
the first edge $(u,n_i)$ on this path.
This step is repeated on the modified graph, until the top-ranked $rec$
changes to a different item.	
\squishend


\textbf{Evaluation Metric.} The metric for assessing the quality of
an explanation is its size, that is, the number of actions in $A^*$
for \prince, and 
the number of edges deleted in $HC$ and $SP$.
\subsection{Results and Insights}
\label{subsec:graph-res}

\begin{table} [t]
	\centering
	\begin{tabular}{c c c c c c c}
		\toprule 
		& \multicolumn{3}{c}{\textbf{Amazon}} & \multicolumn{3}{c}{\textbf{Goodreads}}	\\
		\cmidrule(l){2-4} \cmidrule(l){5-7}
		\textbf{\textit{k}}	& \textbf{\prince} 		& \textbf{\textit{HC}} 	& \textbf{\textit{SP}} 	& \textbf{\prince} 		& \textbf{\textit{HC}}	& \textbf{\textit{SP}}	\\ \toprule
		$3$ 				& $\boldsymbol{5.09}$* 	& $6.87$ 				& $7.57$				& $\boldsymbol{2.05}$* 	& $2.86$ 				& $5.38$				\\
		$5$ 				& $\boldsymbol{3.41}$* 	& $4.62$ 				& $5.01$				& $\boldsymbol{1.66}$* 	& $2.19$ 				& $4.37$				\\
		$10$ 				& $\boldsymbol{2.66}$* 	& $3.66$ 				& $4.15$	 			& $\boldsymbol{1.43}$ 	& $1.45$ 				& $3.28$				\\
		$15$ 				& $\boldsymbol{2.13}$* 	& $3.00$ 				& $3.68$				& $\boldsymbol{1.11}$ 	& $1.12$ 				& $2.90$				\\
		$20$				& $\boldsymbol{1.80}$* 	& $2.39$ 				& $3.28$				& $\boldsymbol{1.11}$	& $1.12$ 				& $2.90$ 				\\ \midrule
	\end{tabular}
	\caption{Average sizes of counterfactual explanations. The best value
		per row in a dataset is in \textbf{bold}. An asterisk (*) indicates
		statistical significance of \prince over
		the closest baseline, under the $1$-tailed paired $t$-test
		at $p < 0.05$.}
	\label{tab:graph-main}
	\vspace*{-0.3cm}
\end{table}

\begin{table} [t]
	\centering
	\begin{tabular}{c c c c c}
		\toprule 
		\multirow{2}{*}{\textbf{Parameter}} & \multicolumn{2}{c}{\textbf{Amazon}} & \multicolumn{2}{c}{\textbf{Goodreads}} \\
		\cmidrule(l){2-3} \cmidrule(l){4-5}
									& \textbf{Pre-comp} & \textbf{Dynamic} 	& \textbf{Pre-comp}	& \textbf{Dynamic} 	\\ \toprule
		$k = 3$ 					& $0.3ms$ 			& $39.1s$ 			& $0.3ms$ 			& $24.1s$ 			\\
		$k = 5$ 					& $0.6ms$ 			& $60.4s$ 			& $0.4ms$ 			& $34.7s$			\\
		$k = 10$ 					& $1.3ms$ 			& $121.6s$ 			& $0.9ms$ 			& $60.7s$ 			\\
		$k = 15$ 					& $2.0ms$ 			& $169.3s$ 			& $1.5ms$ 			& $91.6s$			\\
		$k = 20$					& $2.6ms$ 			& $224.4s$ 			& $2ms$ 			& $118.8s$			\\ \midrule
		$\beta = 0.01$ 				& $0.4ms$ 			& $1.1s$ 			& $0.3ms$ 			& $2.9s$ 			\\
		$\beta = 0.1$ 				& $0.5ms$ 			& $15.5s$ 			& $0.3ms$ 			& $8.9s$ 			\\
		$\beta = 0.3$ 				& $0.5ms$ 			& $17.0s$ 			& $0.4ms$ 			& $12.5s$ 			\\
		$\beta = 0.5$ 				& $0.6ms$ 			& $60.5s$ 			& $0.4ms$ 			& $34.7s$  		\\ \bottomrule
	\end{tabular}
	\caption{Average runtime of \prince,
		when the scores are pre-computed (\textbf{Pre-comp}) and when the scores
		are dynamically computed using
		the reverse push algorithm~\cite{zhang2016approximate}
		(\textbf{Dynamic}).}
	\label{tab:graph-perf}
	\vspace*{-0.3cm}
\end{table}

\begin{table} [!h] 
	\centering
	\resizebox*{\columnwidth}{!}{
		\begin{tabular}{l p{7cm}}
			\toprule
			\textbf{Method}		&	\textbf{Explanation for ``Baby stroller''
									with category ``Baby'' [Amazon]} 													\\	\toprule
			\prince				&	\textbf{Action 1:} You rated highly ``Badger Basket Storage Cubby''
									with category ``Baby'' 																\\
								& 	\textbf{Replacement Item:} ``Google Chromecast HDMI Streaming Media Player''
									with categories ``Home Entertainment''												\\	\midrule
			$HC$				&	\textbf{Action 1:} You rated highly ``Men's hair paste''
									with category ``Beauty'' 															\\
								&	\textbf{Action 2:} You reviewed ``Men's hair paste''
									with category ``Beauty''
									with text ``Good product. Great price.''											\\
								&	\textbf{Action 3:} You rated highly ``Badger Basket Storage Cubby''
									with category ``Baby'' 																\\
								& 	\textbf{Action 4:} You rated highly ``Straw bottle''
									with category ``Baby'' 																\\
								& 	\textbf{Action 5:} You rated highly ``3 Sprouts Storage Caddy''
									with category ``Baby''																\\ 
								&	\textbf{Replacement Item:} ``Bathtub Waste And Overflow Plate''
									with categories ``Home Improvement''												\\	\midrule
			$SP$				&	\textbf{Action 1:} You rated highly ``Men's hair paste''
									with category ``Beauty'' 															\\
								&	\textbf{Action 2:} You rated highly ``Badger Basket Storage Cubby''
									with category ``Baby'' 																\\
								& 	\textbf{Action 3:} You rated highly ``Straw bottle''
								 	with category ``Baby'' 																\\
								& 	\textbf{Action 4:} You rated highly ``3 Sprouts Storage Caddy''
									with category ``Baby'' 																\\
								& 	\textbf{Replacement Item:} ``Google Chromecast HDMI Streaming Media Player''
									with categories ``Home Entertainment''												\\	\toprule \toprule
			\textbf{Method}		&	\textbf{Explanation for ``The Multiversity''
									with categories ``Comics, Historical-fiction, Biography, Mystery'' [Goodreads]} 	\\	\toprule
			\prince 			&	\textbf{Action 1:} You rated highly ``Blackest Night''
									with categories ``Comics, Fantasy, Mystery, Thriller'' 								\\
								& 	\textbf{Action 2: } You rated highly ``Green Lantern''
									with categories ``Comics, Fantasy, Children''										\\ 			
								& 	\textbf{Replacement item:} ``True Patriot: Heroes of the Great White North''
									with categories ``Comics, Fiction''													\\	\midrule			
			$HC$ 				& 	\textbf{Action 1:} You follow User ID $x$ 											\\
								&	\textbf{Action 2:} You rated highly ``Blackest Night''
									with categories ``Comics, Fantasy, Mystery, Thriller'' 								\\
								& 	\textbf{Action 3: } You rated highly ``Green Lantern''
									with categories ``Comics, Fantasy, Children''										\\ 
								& 	\textbf{Replacement item:} ``The Lovecraft Anthology: Volume 2''
									with categories ``Comics, Crime, Fiction''											\\	\midrule			
			$SP$				& 	\textbf{Action 1:}  You follow User ID $x$ 											\\
								& 	\textbf{Action 2:} You rated highly ``Fahrenheit 451''
									with categories ``Fantasy, Young-adult, Fiction''									\\
								& 	\textbf{Action 3:} You rated highly ``Darkly Dreaming Dexter (Dexter, \#1)''
									with categories ``Mystery, Crime, Fantasy'' 										\\ 
								& 	\textbf{And 6 more actions} 														\\
								& 	\textbf{Replacement item:} ``The Lovecraft Anthology: Volume 2''
									with categories ``Comics, Crime, Fiction'' 											\\ \bottomrule
		\end{tabular}}
		\caption{Anecdotal examples of explanations by \prince
			and the counterfactual baselines.}
		\label{tab:anecdotes}
		\vspace*{-0.7cm}
\end{table}

We present our main results in Table~\ref{tab:graph-main} and discuss
insights below. 
These comparisons were performed for different values of the parameter $k$. 
Wherever applicable, statistical significance was tested under
the $1$-tailed paired $t$-test at $p < 0.05$. Anecdotal examples of
explanations by \prince and the baselines
are given in Table~\ref{tab:anecdotes}.
In the Amazon example, we observe that our method produces a topically
coherent explanation, with both the recommendation and the explanation items
in the same category. The $SP$ and $HC$ methods give larger explanations, but with
poorer quality, as the first action in both methods seems unrelated to the recommendation.
In the Goodreads example, both $HC$ and $SP$ yield the same replacement item,
which is different from that of \prince.

\textbf{Approximating \prince is difficult.} 
Explanations generated by \prince are more concise and hence
more user-comprehensible than those by the baselines.
This advantage is quite pronounced; for example, in Amazon, all the 
baselines yield at least one more action in the explanation set on average. 
Note that this translates into unnecessary effort for users who want 
to act upon the explanations.  

\textbf{Explanations shrink with increasing \textit{k}.} The size of
explanations shrinks as the top-$k$ candidate set for choosing
the replacement item is expanded.
For example, the explanation size for \prince on Amazon drops from
$5.09$ at $k = 3$
to $1.80$ at $k = 20$. This is due to the fact
that with a growing candidate set, it becomes
easier to find an item that can 
outrank $rec$. 

\textbf{\prince is efficient.} To generate a counterfactual explanation,
\prince only relies on the scores in the graph configuration  $G \setminus A$ 
(where all the outgoing edges of $u$
are deleted). 
Pre-computing $PPR(n_i, rec|A)$ (for all $n_i \in \mathcal{N}_{out}(u)$), 
\prince could find the explanation for each $(user,rec)$ pair in about 
$1$ millisecond on average (for $k \leq 20$). 
Table~\ref{tab:graph-perf} shows runtimes of \prince for different 
parameters.  
As we can see, the runtime grows
linearly with $k$
in both datasets. This is justified by
Line~\ref{alg:mainstart} in Algorithm~\ref{alg:poly}.
Computing $PPR(n_i, rec|A)$ on-the-fly slows down the algorithm. 
The second and the fourth columns in Table~\ref{tab:graph-perf}
present the runtimes of \prince 
when the scores $PPR(n_i, rec|A)$ are computed using the 
reverse push algorithm for dynamic graphs~\cite{zhang2016approximate}.
Increasing $\beta$ makes the computation slower (experimented at $k=5$).
All experiments were performed on an Intel Xeon server with
$8$ cores @ $3.2$ GHz CPU and $512$ GB main memory. 

\section{User Study}
\label{sec:user}


\textbf{Qualitative survey on usefulness.}
To evaluate the usefulness of counterfactual (action-oriented) explanations,
we conducted a survey with
Amazon Mechanical Turk (AMT) Master
workers (\url{www.mturk.com/help#what_are_masters}).
In this survey, we showed $500$ workers three recommendation items
(``Series Camelot'', ``Pregnancy guide book'', ``Nike backpack'') 
and two different explanations for each.
One explanation was 
limited to only the user's own actions (action-oriented),
and the other was a path
connecting the user to the item (connection-oriented).

We asked the workers three questions:
(i) \textit{Which method do you find more 
useful?}, where $70\%$ chose the action-oriented method;
(ii) \textit{How do you feel about being exposed through explanations
	to others?},
where $\simeq75\%$ expressed a privacy concern either through
complete disapproval or 
through a demand for anonymization;
(iii) \textit{Personally,
	 which type of explanation matters to you more: ``Action-oriented''
	 or ``connection-oriented''?},
 where $61.2\%$ of the workers chose the action-oriented explanations.
We described action-oriented 
explanations as those allowing users to control their recommendation,
while connection-oriented ones reveal connections between the user and item
via other users and items.


\textbf{Quantitative measurement of usefulness.}
In a separate study (conducted only on Amazon data for resource constraints),
we compared \prince to
a path-based explanation~\cite{DBLP:conf/icdm/YangLWH18}
(later referred to as CredPaths).
We used the \textit{credibility} measure
from~\cite{DBLP:conf/icdm/YangLWH18}, scoring paths in descending order of
the product of their edge weights.
We computed
the best path for all $500$ user-item pairs (Sec.~\ref{subsec:graph-setup}).
This resulted in paths of a maximum length of three edges
(four nodes including user and $rec$).
For a fair comparison in terms of cognitive load,
we eliminated all data points
where \prince computed larger counterfactual sets.
This resulted in about $200$ user-item pairs, from where we
sampled exactly $200$. 
As explanations generated by \prince and CredPaths
have a different format of 
presentation (a list of actions vs. a path),
we evaluated each method separately to avoid presentation bias.
For the 
sake of readability, we broke the paths into edges and showed each edge on
a new line. 
Having three AMT Masters for each task, we collected $600 (200 \times 3)$
annotations for \prince and the same number for CredPaths. 

A typical data point
looks like a row in Table~\ref{tab:anecdotes-amt},
that shows representative examples (Goodreads shown only for completeness).
We divided 
the samples into ten HITs (Human Intelligence Tasks, a unit of job on AMT)
with $20$ data points in each HIT. 
For each data point, we showed a recommendation item and its explanation, and
asked users about the usefulness of the explanation on a scale of $1 - 3$
(``Not useful at all'', ``Partially useful'', and ``Completely useful''). 
For this, workers had to 
imagine that they were a user of an e-commerce platform who received the 
recommendations as result of doing some actions on the platform.
Only AMT Master workers were
allowed to provide assessments.

To detect \textit{spammers}, we planted one \textit{honeypot}
in each of the $10$ HITs,
that was a completely impertinent explanation. Subsequently, all
annotations of detected spammers (workers who rated such
irrelevant explanations as ``completely useful'')
were removed (~25\% of all annotations). 


  

\begin{table} [t]
	\centering
	\begin{tabular}{l c c c }
		\toprule 
		\textbf{Method}								& \textbf{Mean} & \textbf{Std. Dev.}	& \textbf{\#Samples}	\\ \toprule
		\prince 									& $1.91$* 		& $0.66$ 				& $200$ 				\\
		CredPaths~\cite{DBLP:conf/icdm/YangLWH18}	& $1.78$ 		& $0.63$ 				& $200$ 				\\ \midrule
		\prince (Size=$1$) 							& $1.87$		& $0.66$ 				& $154$ 				\\ 
		\prince	(Size=$2$)							& $1.88$* 		& $0.70$ 				& $28$					\\
		\prince	(Size=$3$)							& $2.21$*		& $0.52$ 				& $18$					\\ \bottomrule
	\end{tabular}
	\caption{Results from the AMT measurement study on usefulness
		conducted on the Amazon data.
		An asterisk (*) indicates statistical significance of \prince over
		CredPaths ($1$-tailed paired $t$-test at $p < 0.05$).}
	\label{tab:amt-results}
\end{table}

\begin{table} [t] 
	\centering
	\resizebox*{\columnwidth}{!}{
		\begin{tabular}{l p{8 cm}}
			\toprule
			\textbf{Method}		&	\textbf{Explanation for ``Baby stroller''
									with category ``Baby'' [Amazon]} 													\\	\toprule
			\prince				&	\textbf{Action 1:} You rated highly ``Badger Basket Storage Cubby''
									with category ``Baby''																\\
								&	\textbf{Replacement Item:} "Google Chromecast HDMI Streaming Media Player"
									with category ``Home Entertainment"													\\ \midrule			
			CredPaths			&	You rated highly ``Men's hair paste'' with category ``Beauty'' 						\\
								& 	that was rated by ``Some user'' 													\\
								& 	who also rated highly ``Baby stroller'' with category ``Baby''						\\ \toprule
			\textbf{Method}		&	\textbf{Explanation for ``The Multiversity''
									with categories ``Comics, Historical-fiction, Biography, Mystery'' [Goodreads]} 	\\	\toprule
			\prince 			&	\textbf{Action 1:} You rated highly ``Blackest Night''
									with categories ``Comics, Fantasy, Mystery, Thriller'' 								\\
								& 	\textbf{Action 2: } You rated highly ``Green Lantern''
									with categories ``Comics, Fantasy, Children''										\\
								& 	\textbf{Replacement Item:} ``True Patriot: Heroes of the Great White North''
									with categories ``Comics, Fiction, Crime, Fiction'' 								\\ \midrule
			CredPaths			&  	You follow ``Some user''															\\
								&	who has rated highly ``The Multiversity''										
									with categories ``Comics, Historical-fiction, Biography, Mystery'' 					\\ \bottomrule
	\end{tabular}}
	\caption{Explanations from \prince vis-\`{a}-vis CredPaths~\cite{DBLP:conf/icdm/YangLWH18}.}
	\label{tab:anecdotes-amt}
\end{table}

\begin{table} [t]
	\centering
	\resizebox*{\columnwidth}{!}{
		\begin{tabular}{p{8.5cm}}
			\toprule
			Based on multiple actions explained simply and clearly. [\prince$\!$] 							\\ \midrule
			The recommendation is for a home plumbing item, but
			the action rated a glue. [\prince$\!$] 		\\ \midrule
			The explanation is complete as it goes into full details of how
			to use the product,
			which is in alignment of my review and useful to me. [CredPaths] 								\\ \midrule
			It's weird to be given recommendations based on
			other people. [CredPaths] 						\\ \bottomrule
	\end{tabular}}
	\caption{Turkers' comments on their score justifications.}
	\label{tab:comments}
\end{table}

Table~\ref{tab:amt-results} shows the results of
our user study.
It gives average scores and standard deviations,
and it indicates statistical significance of pairwise comparisons
with an asterisk.
\prince clearly obtains higher usefulness ratings from the AMT judges,
on average. Krippendorff's alpha~\cite{krippendorff2018content}
for \prince and CredPaths were found to be
$\simeq 0.5$ and $\simeq 0.3$ respectively, showing moderate
to fair inter-annotator agreement. 
The superiority of \prince also holds for slices of samples where 
\prince generated explanations of size $1, 2$ and $3$. 
We also asked Turkers to provide \textit{succinct justifications}
for their scores
on each data point.
Table~\ref{tab:comments} shows some typical comments, where methods
for generating explanations are in brackets.

\section{Related Work}
\label{sec:related}

Foundational work 
on explainability for collaborative-filtering-based recommenders
was done by Herlocker et al.~\cite{herlocker2000explaining}. Over time, 
generating
explanations (like~\cite{DBLP:conf/icdm/YangLWH18}) has become
tightly coupled with building 
systems that are geared for producing more transparent 
recommendations (like~\cite{balog2019transparent}). 
For broad surveys, see
\cite{zhang2018explainable,tintarev2007survey}.
%
With methods
using matrix or tensor
factorization~\cite{zhang2014explicit,chen2016learning,wang2018explainable},
the goal has been to make latent factors more tangible.
Recently, interpretable neural models have become popular, especially for 
text~\cite{seo2017interpretable,chen2018neural,chen2018sequential} and
images~\cite{chen2019personalized}, where the attention mechanism
over words, reviews, items, or zones in images
has been
vital for interpretability. Efforts have also been made on generating readable 
explanations using models like LSTMs~\cite{li2017neural}
or GANs~\cite{lu2018like}.

%
Representing users, items, categories and reviews as
a knowledge graph or a heterogeneous information network (HIN) has 
become popular, where explanations
take the form of paths between the user and an item. This paradigm
comprises a variety of mechanisms:
learning path embeddings~\cite{ai2018learning,wang2018explainable},
propagating user preferences~\cite{wang2018ripplenet},
learning and reasoning with explainable 
rules~\cite{ma2019jointly,xian2019reinforcement},
and ranking user-item connections~\cite{DBLP:conf/icdm/YangLWH18,ghazimatin:19}. 
In this work,
we choose the recent approach in~\cite{DBLP:conf/icdm/YangLWH18} as
a representative for the family of path-based recommenders to compare
\prince with.
Finally, post-hoc or model-agnostic rationalizations for black-box models have 
attracted interest.
Approaches
include association rule mining~\cite{peake2018explanation},
supervised ranking of user-item relationships~\cite{ghazimatin:19},
and reinforcement learning~\cite{wang2018reinforcement}.
%
%

Random walks over HIN's have been pursued by
a suite of works, including \cite{DBLP:reference/rsh/DesrosiersK11,DBLP:conf/www/CooperLRS14,DBLP:conf/recsys/ChristoffelPNB15,DBLP:conf/www/EksombatchaiJLL18,jiang2018recommendation}.
In a nutshell, the Personalized PageRank (PPR) of an item node in the HIN 
is used as a ranking criterion for recommendations.
%
\cite{nikolakopoulos2019recwalk} introduced 
the RecWalk method, proposing a random walk
with a nearly uncoupled Markov chain.
%
Our work uses this framework.
As far as we know, we are the 
first to study the problem of computing minimum subsets of edge removals
(user actions) to change the top-ranked node in a counterfactual setup.
Prior research on dynamic graphs, 
such as \cite{csaji2014pagerank,kang2018aurora},
 has addressed related issues, but not this very problem.
A separate line of research focuses on the efficient computation of PPR.
Approximate algorithms include power iteration~\cite{page1999pagerank},
local push~\cite{andersen2006local, andersen2007local, zhang2016approximate}
and Monte Carlo methods~\cite{avrachenkov2007monte, bahmani2010fast}. 

\vspace*{-0.1cm}
\section{Conclusions and Future Work}
\label{sec:confut}

This work explored a new paradigm of action-based explanations in
graph
recommenders, with the goal of identifying minimum sets of user actions
with the counterfactual property that their absence would change the
top-ranked recommendation to a different item. 
In contrast to prior works on (largely path-based) recommender explanations, this 
approach offers two advantages:
(i) explanations are \textit{concise, scrutable}, and \textit{actionable},
as they are \textit{minimal} sets derived using
a \textit{counterfactual} setup over
a user's \textit{own} purchases, ratings and reviews; and
(ii) explanations do not expose any information about other users, thus avoiding
privacy breaches by design.

%

The proposed \prince method implements these principles using
random walks for Personalized PageRank scores
as a recommender model.
We presented an efficient computation and correctness proof for
computing counterfactual explanations, despite the potentially exponential
search space of user-action subsets.
Extensive experiments
on large real-life data from Amazon and Goodreads showed that 
simpler heuristics  fail to find the best explanations, whereas \prince
can guarantee optimality.
Studies with AMT Masters showed the superiority of
\prince over baselines in terms of explanation usefulness.
%

\vspace*{-0.3cm}
\section*{Acknowledgements}
 
This work was partly supported by the ERC
Synergy Grant 610150 (imPACT) and the DFG Collaborative Research 
Center 1223. We would like to thank Simon Razniewski from
the MPI for Informatics
for his insightful comments on the manuscript.


\bibliographystyle{ACM-Reference-Format}
\bibliography{2020-arxiv-fp-prince}


\begin{thebibliography}{60}


\ifx \showCODEN    \undefined \def \showCODEN     #1{\unskip}     \fi
\ifx \showDOI      \undefined \def \showDOI       #1{#1}\fi
\ifx \showISBNx    \undefined \def \showISBNx     #1{\unskip}     \fi
\ifx \showISBNxiii \undefined \def \showISBNxiii  #1{\unskip}     \fi
\ifx \showISSN     \undefined \def \showISSN      #1{\unskip}     \fi
\ifx \showLCCN     \undefined \def \showLCCN      #1{\unskip}     \fi
\ifx \shownote     \undefined \def \shownote      #1{#1}          \fi
\ifx \showarticletitle \undefined \def \showarticletitle #1{#1}   \fi
\ifx \showURL      \undefined \def \showURL       {\relax}        \fi
\providecommand\bibfield[2]{#2}
\providecommand\bibinfo[2]{#2}
\providecommand\natexlab[1]{#1}
\providecommand\showeprint[2][]{arXiv:#2}

\bibitem[\protect\citeauthoryear{Ai, Azizi, Chen, and Zhang}{Ai
  et~al\mbox{.}}{2018}]%
        {ai2018learning}
\bibfield{author}{\bibinfo{person}{Qingyao Ai}, \bibinfo{person}{Vahid Azizi},
  \bibinfo{person}{Xu Chen}, {and} \bibinfo{person}{Yongfeng Zhang}.}
  \bibinfo{year}{2018}\natexlab{}.
\newblock \showarticletitle{Learning heterogeneous knowledge base embeddings
  for explainable recommendation}.
\newblock \bibinfo{journal}{\emph{Algorithms}} \bibinfo{volume}{11},
  \bibinfo{number}{9} (\bibinfo{year}{2018}).
\newblock


\bibitem[\protect\citeauthoryear{Andersen, Borgs, Chayes, Hopcraft, Mirrokni,
  and Teng}{Andersen et~al\mbox{.}}{2007}]%
        {andersen2007local}
\bibfield{author}{\bibinfo{person}{Reid Andersen}, \bibinfo{person}{Christian
  Borgs}, \bibinfo{person}{Jennifer Chayes}, \bibinfo{person}{John Hopcraft},
  \bibinfo{person}{Vahab~S Mirrokni}, {and} \bibinfo{person}{Shang-Hua Teng}.}
  \bibinfo{year}{2007}\natexlab{}.
\newblock \showarticletitle{{Local computation of PageRank contributions}}. In
  \bibinfo{booktitle}{\emph{WAW}}.
\newblock


\bibitem[\protect\citeauthoryear{Andersen, Chung, and Lang}{Andersen
  et~al\mbox{.}}{2006}]%
        {andersen2006local}
\bibfield{author}{\bibinfo{person}{Reid Andersen}, \bibinfo{person}{Fan Chung},
  {and} \bibinfo{person}{Kevin Lang}.} \bibinfo{year}{2006}\natexlab{}.
\newblock \showarticletitle{{Local graph partitioning using Pagerank vectors}}.
  In \bibinfo{booktitle}{\emph{FOCS}}.
\newblock


\bibitem[\protect\citeauthoryear{Avrachenkov, Litvak, Nemirovsky, and
  Osipova}{Avrachenkov et~al\mbox{.}}{2007}]%
        {avrachenkov2007monte}
\bibfield{author}{\bibinfo{person}{Konstantin Avrachenkov},
  \bibinfo{person}{Nelly Litvak}, \bibinfo{person}{Danil Nemirovsky}, {and}
  \bibinfo{person}{Natalia Osipova}.} \bibinfo{year}{2007}\natexlab{}.
\newblock \showarticletitle{Monte Carlo methods in PageRank computation: When
  one iteration is sufficient}.
\newblock \bibinfo{journal}{\emph{SIAM J. Numer. Anal.}} \bibinfo{volume}{45},
  \bibinfo{number}{2} (\bibinfo{year}{2007}).
\newblock


\bibitem[\protect\citeauthoryear{Bahmani, Chowdhury, and Goel}{Bahmani
  et~al\mbox{.}}{2010}]%
        {bahmani2010fast}
\bibfield{author}{\bibinfo{person}{Bahman Bahmani}, \bibinfo{person}{Abdur
  Chowdhury}, {and} \bibinfo{person}{Ashish Goel}.}
  \bibinfo{year}{2010}\natexlab{}.
\newblock \showarticletitle{{Fast incremental and personalized PageRank}}. In
  \bibinfo{booktitle}{\emph{VLDB}}.
\newblock


\bibitem[\protect\citeauthoryear{Balog, Radlinski, and Arakelyan}{Balog
  et~al\mbox{.}}{2019}]%
        {balog2019transparent}
\bibfield{author}{\bibinfo{person}{Krisztian Balog}, \bibinfo{person}{Filip
  Radlinski}, {and} \bibinfo{person}{Shushan Arakelyan}.}
  \bibinfo{year}{2019}\natexlab{}.
\newblock \showarticletitle{Transparent, Scrutable and Explainable User Models
  for Personalized Recommendation}. In \bibinfo{booktitle}{\emph{SIGIR}}.
\newblock


\bibitem[\protect\citeauthoryear{Brin and Page}{Brin and Page}{1998}]%
        {brin1998anatomy}
\bibfield{author}{\bibinfo{person}{Sergey Brin} {and} \bibinfo{person}{Lawrence
  Page}.} \bibinfo{year}{1998}\natexlab{}.
\newblock \showarticletitle{The anatomy of a large-scale hypertextual web
  search engine}.
\newblock \bibinfo{journal}{\emph{Computer networks and ISDN systems}}
  \bibinfo{volume}{30}, \bibinfo{number}{1-7} (\bibinfo{year}{1998}).
\newblock


\bibitem[\protect\citeauthoryear{Cer, Yang, Kong, Hua, Limtiaco, St.~John,
  Constant, Guajardo-Cespedes, Yuan, Tar, Strope, and Kurzweil}{Cer
  et~al\mbox{.}}{2018}]%
        {cer2018universal}
\bibfield{author}{\bibinfo{person}{Daniel Cer}, \bibinfo{person}{Yinfei Yang},
  \bibinfo{person}{Sheng-yi Kong}, \bibinfo{person}{Nan Hua},
  \bibinfo{person}{Nicole Limtiaco}, \bibinfo{person}{Rhomni St.~John},
  \bibinfo{person}{Noah Constant}, \bibinfo{person}{Mario Guajardo-Cespedes},
  \bibinfo{person}{Steve Yuan}, \bibinfo{person}{Chris Tar},
  \bibinfo{person}{Brian Strope}, {and} \bibinfo{person}{Ray Kurzweil}.}
  \bibinfo{year}{2018}\natexlab{}.
\newblock \showarticletitle{Universal Sentence Encoder for {E}nglish}. In
  \bibinfo{booktitle}{\emph{EMNLP}}.
\newblock


\bibitem[\protect\citeauthoryear{Chen, Zhang, Liu, and Ma}{Chen
  et~al\mbox{.}}{2018b}]%
        {chen2018neural}
\bibfield{author}{\bibinfo{person}{Chong Chen}, \bibinfo{person}{Min Zhang},
  \bibinfo{person}{Yiqun Liu}, {and} \bibinfo{person}{Shaoping Ma}.}
  \bibinfo{year}{2018}\natexlab{b}.
\newblock \showarticletitle{Neural attentional rating regression with
  review-level explanations}. In \bibinfo{booktitle}{\emph{WWW}}.
\newblock


\bibitem[\protect\citeauthoryear{Chen, Fraiberger, Moakler, and Provost}{Chen
  et~al\mbox{.}}{2017}]%
        {chen2017enhancing}
\bibfield{author}{\bibinfo{person}{Daizhuo Chen}, \bibinfo{person}{Samuel~P.
  Fraiberger}, \bibinfo{person}{Robert Moakler}, {and} \bibinfo{person}{Foster
  Provost}.} \bibinfo{year}{2017}\natexlab{}.
\newblock \showarticletitle{Enhancing transparency and control when drawing
  data-driven inferences about individuals}.
\newblock \bibinfo{journal}{\emph{Big data}} \bibinfo{volume}{5},
  \bibinfo{number}{3} (\bibinfo{year}{2017}).
\newblock


\bibitem[\protect\citeauthoryear{Chen, Chen, Xu, Zhang, Cao, Qin, and Zha}{Chen
  et~al\mbox{.}}{2019}]%
        {chen2019personalized}
\bibfield{author}{\bibinfo{person}{Xu Chen}, \bibinfo{person}{Hanxiong Chen},
  \bibinfo{person}{Hongteng Xu}, \bibinfo{person}{Yongfeng Zhang},
  \bibinfo{person}{Yixin Cao}, \bibinfo{person}{Zheng Qin}, {and}
  \bibinfo{person}{Hongyuan Zha}.} \bibinfo{year}{2019}\natexlab{}.
\newblock \showarticletitle{{Personalized Fashion Recommendation with Visual
  Explanations based on Multimodal Attention Network: Towards Visually
  Explainable Recommendation}}. In \bibinfo{booktitle}{\emph{SIGIR}}.
\newblock


\bibitem[\protect\citeauthoryear{Chen, Qin, Zhang, and Xu}{Chen
  et~al\mbox{.}}{2016}]%
        {chen2016learning}
\bibfield{author}{\bibinfo{person}{Xu Chen}, \bibinfo{person}{Zheng Qin},
  \bibinfo{person}{Yongfeng Zhang}, {and} \bibinfo{person}{Tao Xu}.}
  \bibinfo{year}{2016}\natexlab{}.
\newblock \showarticletitle{Learning to Rank Features for Recommendation over
  Multiple Categories}. In \bibinfo{booktitle}{\emph{SIGIR}}.
\newblock


\bibitem[\protect\citeauthoryear{Chen, Xu, Zhang, Tang, Cao, Qin, and Zha}{Chen
  et~al\mbox{.}}{2018a}]%
        {chen2018sequential}
\bibfield{author}{\bibinfo{person}{Xu Chen}, \bibinfo{person}{Hongteng Xu},
  \bibinfo{person}{Yongfeng Zhang}, \bibinfo{person}{Jiaxi Tang},
  \bibinfo{person}{Yixin Cao}, \bibinfo{person}{Zheng Qin}, {and}
  \bibinfo{person}{Hongyuan Zha}.} \bibinfo{year}{2018}\natexlab{a}.
\newblock \showarticletitle{Sequential recommendation with user memory
  networks}. In \bibinfo{booktitle}{\emph{WSDM}}.
\newblock


\bibitem[\protect\citeauthoryear{Christoffel, Paudel, Newell, and
  Bernstein}{Christoffel et~al\mbox{.}}{2015}]%
        {DBLP:conf/recsys/ChristoffelPNB15}
\bibfield{author}{\bibinfo{person}{Fabian Christoffel}, \bibinfo{person}{Bibek
  Paudel}, \bibinfo{person}{Chris Newell}, {and} \bibinfo{person}{Abraham
  Bernstein}.} \bibinfo{year}{2015}\natexlab{}.
\newblock \showarticletitle{{Blockbusters and Wallflowers: Accurate, Diverse,
  and Scalable Recommendations with Random Walks}}. In
  \bibinfo{booktitle}{\emph{RecSys}}.
\newblock


\bibitem[\protect\citeauthoryear{Cooper, Lee, Radzik, and Siantos}{Cooper
  et~al\mbox{.}}{2014}]%
        {DBLP:conf/www/CooperLRS14}
\bibfield{author}{\bibinfo{person}{Colin Cooper}, \bibinfo{person}{Sang{-}Hyuk
  Lee}, \bibinfo{person}{Tomasz Radzik}, {and} \bibinfo{person}{Yiannis
  Siantos}.} \bibinfo{year}{2014}\natexlab{}.
\newblock \showarticletitle{{Random walks in recommender systems: Exact
  computation and simulations}}. In \bibinfo{booktitle}{\emph{WWW}}.
\newblock


\bibitem[\protect\citeauthoryear{Cs{\'a}ji, Jungers, and Blondel}{Cs{\'a}ji
  et~al\mbox{.}}{2014}]%
        {csaji2014pagerank}
\bibfield{author}{\bibinfo{person}{Bal{\'a}zs~Csan{\'a}d Cs{\'a}ji},
  \bibinfo{person}{Rapha{\"e}l~M. Jungers}, {and} \bibinfo{person}{Vincent
  Blondel}.} \bibinfo{year}{2014}\natexlab{}.
\newblock \showarticletitle{PageRank optimization by edge selection}.
\newblock \bibinfo{journal}{\emph{Discrete Applied Mathematics}}
  \bibinfo{volume}{169} (\bibinfo{year}{2014}).
\newblock


\bibitem[\protect\citeauthoryear{Desrosiers and Karypis}{Desrosiers and
  Karypis}{2011}]%
        {DBLP:reference/rsh/DesrosiersK11}
\bibfield{author}{\bibinfo{person}{Christian Desrosiers} {and}
  \bibinfo{person}{George Karypis}.} \bibinfo{year}{2011}\natexlab{}.
\newblock \showarticletitle{A Comprehensive Survey of Neighborhood-based
  Recommendation Methods}.
\newblock In \bibinfo{booktitle}{\emph{Recommender Systems Handbook}}.
\newblock


\bibitem[\protect\citeauthoryear{Eksombatchai, Jindal, Liu, Liu, Sharma,
  Sugnet, Ulrich, and Leskovec}{Eksombatchai et~al\mbox{.}}{2018}]%
        {DBLP:conf/www/EksombatchaiJLL18}
\bibfield{author}{\bibinfo{person}{Chantat Eksombatchai},
  \bibinfo{person}{Pranav Jindal}, \bibinfo{person}{Jerry~Zitao Liu},
  \bibinfo{person}{Yuchen Liu}, \bibinfo{person}{Rahul Sharma},
  \bibinfo{person}{Charles Sugnet}, \bibinfo{person}{Mark Ulrich}, {and}
  \bibinfo{person}{Jure Leskovec}.} \bibinfo{year}{2018}\natexlab{}.
\newblock \showarticletitle{Pixie: {A} System for Recommending 3+ Billion Items
  to 200+ Million Users in Real-Time}. In \bibinfo{booktitle}{\emph{WWW}}.
\newblock


\bibitem[\protect\citeauthoryear{Ghazimatin, Saha~Roy, and Weikum}{Ghazimatin
  et~al\mbox{.}}{2019}]%
        {ghazimatin:19}
\bibfield{author}{\bibinfo{person}{Azin Ghazimatin}, \bibinfo{person}{Rishiraj
  Saha~Roy}, {and} \bibinfo{person}{Gerhard Weikum}.}
  \bibinfo{year}{2019}\natexlab{}.
\newblock \showarticletitle{{FAIRY: A Framework for Understanding Relationships
  between Users' Actions and their Social Feeds}}. In
  \bibinfo{booktitle}{\emph{WSDM}}.
\newblock


\bibitem[\protect\citeauthoryear{Haveliwala}{Haveliwala}{2003}]%
        {haveliwala2003topic}
\bibfield{author}{\bibinfo{person}{Taher~H. Haveliwala}.}
  \bibinfo{year}{2003}\natexlab{}.
\newblock \showarticletitle{{Topic-sensitive Pagerank: A context-sensitive
  ranking algorithm for Web search}}.
\newblock \bibinfo{journal}{\emph{TKDE}} \bibinfo{volume}{15},
  \bibinfo{number}{4} (\bibinfo{year}{2003}).
\newblock


\bibitem[\protect\citeauthoryear{Herlocker, Konstan, and Riedl}{Herlocker
  et~al\mbox{.}}{2000}]%
        {herlocker2000explaining}
\bibfield{author}{\bibinfo{person}{Jonathan~L. Herlocker},
  \bibinfo{person}{Joseph~A. Konstan}, {and} \bibinfo{person}{John Riedl}.}
  \bibinfo{year}{2000}\natexlab{}.
\newblock \showarticletitle{Explaining Collaborative Filtering
  Recommendations}. In \bibinfo{booktitle}{\emph{CSCW}}.
\newblock


\bibitem[\protect\citeauthoryear{Jamali and Ester}{Jamali and Ester}{2009}]%
        {DBLP:conf/kdd/JamaliE09}
\bibfield{author}{\bibinfo{person}{Mohsen Jamali} {and} \bibinfo{person}{Martin
  Ester}.} \bibinfo{year}{2009}\natexlab{}.
\newblock \showarticletitle{\emph{TrustWalker: A} random walk model for
  combining trust-based and item-based recommendation}. In
  \bibinfo{booktitle}{\emph{KDD}}.
\newblock


\bibitem[\protect\citeauthoryear{Jeh and Widom}{Jeh and Widom}{2003}]%
        {jeh2003scaling}
\bibfield{author}{\bibinfo{person}{Glen Jeh} {and} \bibinfo{person}{Jennifer
  Widom}.} \bibinfo{year}{2003}\natexlab{}.
\newblock \showarticletitle{{Scaling personalized Web search}}. In
  \bibinfo{booktitle}{\emph{WWW}}.
\newblock


\bibitem[\protect\citeauthoryear{Jiang, Liu, Fu, Wu, and Zhang}{Jiang
  et~al\mbox{.}}{2018}]%
        {jiang2018recommendation}
\bibfield{author}{\bibinfo{person}{Zhengshen Jiang}, \bibinfo{person}{Hongzhi
  Liu}, \bibinfo{person}{Bin Fu}, \bibinfo{person}{Zhonghai Wu}, {and}
  \bibinfo{person}{Tao Zhang}.} \bibinfo{year}{2018}\natexlab{}.
\newblock \showarticletitle{Recommendation in heterogeneous information
  networks based on generalized random walk model and Bayesian personalized
  ranking}. In \bibinfo{booktitle}{\emph{WSDM}}.
\newblock


\bibitem[\protect\citeauthoryear{Kang, Wang, Cao, Xia, Fan, and Tong}{Kang
  et~al\mbox{.}}{2018}]%
        {kang2018aurora}
\bibfield{author}{\bibinfo{person}{Jian Kang}, \bibinfo{person}{Meijia Wang},
  \bibinfo{person}{Nan Cao}, \bibinfo{person}{Yinglong Xia},
  \bibinfo{person}{Wei Fan}, {and} \bibinfo{person}{Hanghang Tong}.}
  \bibinfo{year}{2018}\natexlab{}.
\newblock \showarticletitle{{AURORA: Auditing PageRank} on Large Graphs}. In
  \bibinfo{booktitle}{\emph{Big Data}}.
\newblock


\bibitem[\protect\citeauthoryear{Koren, Bell, and Volinsky}{Koren
  et~al\mbox{.}}{2009}]%
        {koren2009matrix}
\bibfield{author}{\bibinfo{person}{Yehuda Koren}, \bibinfo{person}{Robert
  Bell}, {and} \bibinfo{person}{Chris Volinsky}.}
  \bibinfo{year}{2009}\natexlab{}.
\newblock \showarticletitle{Matrix factorization techniques for recommender
  systems}.
\newblock \bibinfo{journal}{\emph{Computer}} \bibinfo{number}{8}
  (\bibinfo{year}{2009}).
\newblock


\bibitem[\protect\citeauthoryear{Kouki, Schaffer, Pujara, O'Donovan, and
  Getoor}{Kouki et~al\mbox{.}}{2019}]%
        {kouki2019personalized}
\bibfield{author}{\bibinfo{person}{Pigi Kouki}, \bibinfo{person}{James
  Schaffer}, \bibinfo{person}{Jay Pujara}, \bibinfo{person}{John O'Donovan},
  {and} \bibinfo{person}{Lise Getoor}.} \bibinfo{year}{2019}\natexlab{}.
\newblock \showarticletitle{Personalized explanations for hybrid recommender
  systems}. In \bibinfo{booktitle}{\emph{IUI}}.
\newblock


\bibitem[\protect\citeauthoryear{Krippendorff}{Krippendorff}{2018}]%
        {krippendorff2018content}
\bibfield{author}{\bibinfo{person}{Klaus Krippendorff}.}
  \bibinfo{year}{2018}\natexlab{}.
\newblock \bibinfo{booktitle}{\emph{Content analysis: An introduction to its
  methodology}}.
\newblock \bibinfo{publisher}{Sage}.
\newblock


\bibitem[\protect\citeauthoryear{Kunkel, Donkers, Michael, Barbu, and
  Ziegler}{Kunkel et~al\mbox{.}}{2019}]%
        {kunkel2019let}
\bibfield{author}{\bibinfo{person}{Johannes Kunkel}, \bibinfo{person}{Tim
  Donkers}, \bibinfo{person}{Lisa Michael}, \bibinfo{person}{Catalin-Mihai
  Barbu}, {and} \bibinfo{person}{J{\"u}rgen Ziegler}.}
  \bibinfo{year}{2019}\natexlab{}.
\newblock \showarticletitle{{Let Me Explain: Impact of Personal and Impersonal
  Explanations on Trust in Recommender Systems}}. In
  \bibinfo{booktitle}{\emph{CHI}}.
\newblock


\bibitem[\protect\citeauthoryear{Li, Wang, Ren, Bing, and Lam}{Li
  et~al\mbox{.}}{2017}]%
        {li2017neural}
\bibfield{author}{\bibinfo{person}{Piji Li}, \bibinfo{person}{Zihao Wang},
  \bibinfo{person}{Zhaochun Ren}, \bibinfo{person}{Lidong Bing}, {and}
  \bibinfo{person}{Wai Lam}.} \bibinfo{year}{2017}\natexlab{}.
\newblock \showarticletitle{Neural rating regression with abstractive tips
  generation for recommendation}. In \bibinfo{booktitle}{\emph{SIGIR}}.
\newblock


\bibitem[\protect\citeauthoryear{Lu, Dong, and Smyth}{Lu et~al\mbox{.}}{2018}]%
        {lu2018like}
\bibfield{author}{\bibinfo{person}{Yichao Lu}, \bibinfo{person}{Ruihai Dong},
  {and} \bibinfo{person}{Barry Smyth}.} \bibinfo{year}{2018}\natexlab{}.
\newblock \showarticletitle{{Why I like it: Multi-task learning for
  recommendation and explanation}}. In \bibinfo{booktitle}{\emph{RecSys}}.
\newblock


\bibitem[\protect\citeauthoryear{Ma, Zhang, Cao, Jin, Wang, Liu, Ma, and
  Ren}{Ma et~al\mbox{.}}{2019}]%
        {ma2019jointly}
\bibfield{author}{\bibinfo{person}{Weizhi Ma}, \bibinfo{person}{Min Zhang},
  \bibinfo{person}{Yue Cao}, \bibinfo{person}{Woojeong Jin},
  \bibinfo{person}{Chenyang Wang}, \bibinfo{person}{Yiqun Liu},
  \bibinfo{person}{Shaoping Ma}, {and} \bibinfo{person}{Xiang Ren}.}
  \bibinfo{year}{2019}\natexlab{}.
\newblock \showarticletitle{Jointly Learning Explainable Rules for
  Recommendation with Knowledge Graph}. In \bibinfo{booktitle}{\emph{WWW}}.
\newblock


\bibitem[\protect\citeauthoryear{Machanavajjhala, Korolova, and
  Sarma}{Machanavajjhala et~al\mbox{.}}{2011}]%
        {machanavajjhala2011personalized}
\bibfield{author}{\bibinfo{person}{Ashwin Machanavajjhala},
  \bibinfo{person}{Aleksandra Korolova}, {and} \bibinfo{person}{Atish~Das
  Sarma}.} \bibinfo{year}{2011}\natexlab{}.
\newblock \showarticletitle{{Personalized social recommendations: Accurate or
  private?}}. In \bibinfo{booktitle}{\emph{VLDB}}.
\newblock


\bibitem[\protect\citeauthoryear{Martens and Provost}{Martens and
  Provost}{2014}]%
        {martens2014explaining}
\bibfield{author}{\bibinfo{person}{David Martens} {and} \bibinfo{person}{Foster
  Provost}.} \bibinfo{year}{2014}\natexlab{}.
\newblock \showarticletitle{Explaining Data-Driven Document Classifications}.
\newblock \bibinfo{journal}{\emph{MIS Quarterly}} \bibinfo{volume}{38},
  \bibinfo{number}{1} (\bibinfo{year}{2014}).
\newblock


\bibitem[\protect\citeauthoryear{Miller, Weber, Aha, and Magazzeni}{Miller
  et~al\mbox{.}}{2019}]%
        {miller2019xai}
\bibfield{author}{\bibinfo{person}{Tim Miller}, \bibinfo{person}{Rosina Weber},
  \bibinfo{person}{David Aha}, {and} \bibinfo{person}{Daniele Magazzeni}.}
  \bibinfo{year}{2019}\natexlab{}.
\newblock \showarticletitle{IJCAI 2019 Workshop on Explainable AI (XAI)}.
\newblock


\bibitem[\protect\citeauthoryear{Moeyersoms, d'Alessandro, Provost, and
  Martens}{Moeyersoms et~al\mbox{.}}{2016}]%
        {moeyersoms2016explaining}
\bibfield{author}{\bibinfo{person}{Julie Moeyersoms}, \bibinfo{person}{Brian
  d'Alessandro}, \bibinfo{person}{Foster Provost}, {and} \bibinfo{person}{David
  Martens}.} \bibinfo{year}{2016}\natexlab{}.
\newblock \showarticletitle{Explaining classification models built on
  high-dimensional sparse data}.
\newblock \bibinfo{journal}{\emph{arXiv preprint arXiv:1607.06280}}
  (\bibinfo{year}{2016}).
\newblock


\bibitem[\protect\citeauthoryear{Nikolakopoulos and Karypis}{Nikolakopoulos and
  Karypis}{2019}]%
        {nikolakopoulos2019recwalk}
\bibfield{author}{\bibinfo{person}{Athanasios~N. Nikolakopoulos} {and}
  \bibinfo{person}{George Karypis}.} \bibinfo{year}{2019}\natexlab{}.
\newblock \showarticletitle{{Recwalk: Nearly} uncoupled random walks for top-n
  recommendation}. In \bibinfo{booktitle}{\emph{WSDM}}.
\newblock


\bibitem[\protect\citeauthoryear{Page, Brin, Motwani, and Winograd}{Page
  et~al\mbox{.}}{1999}]%
        {page1999pagerank}
\bibfield{author}{\bibinfo{person}{Lawrence Page}, \bibinfo{person}{Sergey
  Brin}, \bibinfo{person}{Rajeev Motwani}, {and} \bibinfo{person}{Terry
  Winograd}.} \bibinfo{year}{1999}\natexlab{}.
\newblock \bibinfo{booktitle}{\emph{{The PageRank citation ranking: Bringing
  order to the Web}}}.
\newblock \bibinfo{type}{{T}echnical {R}eport}. \bibinfo{institution}{Stanford
  InfoLab}.
\newblock


\bibitem[\protect\citeauthoryear{Peake and Wang}{Peake and Wang}{2018}]%
        {peake2018explanation}
\bibfield{author}{\bibinfo{person}{Georgina Peake} {and} \bibinfo{person}{Jun
  Wang}.} \bibinfo{year}{2018}\natexlab{}.
\newblock \showarticletitle{{Explanation mining: Post hoc interpretability of
  latent factor models for recommendation systems}}. In
  \bibinfo{booktitle}{\emph{KDD}}.
\newblock


\bibitem[\protect\citeauthoryear{Ribeiro, Singh, and Guestrin}{Ribeiro
  et~al\mbox{.}}{2016}]%
        {ribeiro2016should}
\bibfield{author}{\bibinfo{person}{Marco~Tulio Ribeiro},
  \bibinfo{person}{Sameer Singh}, {and} \bibinfo{person}{Carlos Guestrin}.}
  \bibinfo{year}{2016}\natexlab{}.
\newblock \showarticletitle{{Why should I trust you?: Explaining the
  predictions of any classifier}}. In \bibinfo{booktitle}{\emph{KDD}}.
\newblock


\bibitem[\protect\citeauthoryear{Rimchala, Doshi, Zhu, Chang, Hoh, Peuter,
  Lador, and Dasgupta}{Rimchala et~al\mbox{.}}{2019}]%
        {rimchala2019kdd}
\bibfield{author}{\bibinfo{person}{Joy Rimchala}, \bibinfo{person}{Jineet
  Doshi}, \bibinfo{person}{Qiang Zhu}, \bibinfo{person}{Diane Chang},
  \bibinfo{person}{Nick Hoh}, \bibinfo{person}{Conrad~De Peuter},
  \bibinfo{person}{Shir~Meir Lador}, {and} \bibinfo{person}{Sambarta
  Dasgupta}.} \bibinfo{year}{2019}\natexlab{}.
\newblock \showarticletitle{{KDD Workshop on Explainable AI for Fairness,
  Accountability, and Transparency}}.
\newblock


\bibitem[\protect\citeauthoryear{Seo, Huang, Yang, and Liu}{Seo
  et~al\mbox{.}}{2017}]%
        {seo2017interpretable}
\bibfield{author}{\bibinfo{person}{Sungyong Seo}, \bibinfo{person}{Jing Huang},
  \bibinfo{person}{Hao Yang}, {and} \bibinfo{person}{Yan Liu}.}
  \bibinfo{year}{2017}\natexlab{}.
\newblock \showarticletitle{Interpretable convolutional neural networks with
  dual local and global attention for review rating prediction}. In
  \bibinfo{booktitle}{\emph{RecSys}}.
\newblock


\bibitem[\protect\citeauthoryear{Shi, Li, Zhang, Sun, and Yu}{Shi
  et~al\mbox{.}}{2017}]%
        {DBLP:journals/tkde/ShiLZSY17}
\bibfield{author}{\bibinfo{person}{Chuan Shi}, \bibinfo{person}{Yitong Li},
  \bibinfo{person}{Jiawei Zhang}, \bibinfo{person}{Yizhou Sun}, {and}
  \bibinfo{person}{Philip~S. Yu}.} \bibinfo{year}{2017}\natexlab{}.
\newblock \showarticletitle{A Survey of Heterogeneous Information Network
  Analysis}.
\newblock \bibinfo{journal}{\emph{TKDE}} \bibinfo{volume}{29},
  \bibinfo{number}{1} (\bibinfo{year}{2017}).
\newblock


\bibitem[\protect\citeauthoryear{Shi, Zhang, Luo, Yu, Yue, and Wu}{Shi
  et~al\mbox{.}}{2015}]%
        {shi2015semantic}
\bibfield{author}{\bibinfo{person}{Chuan Shi}, \bibinfo{person}{Zhiqiang
  Zhang}, \bibinfo{person}{Ping Luo}, \bibinfo{person}{Philip~S Yu},
  \bibinfo{person}{Yading Yue}, {and} \bibinfo{person}{Bin Wu}.}
  \bibinfo{year}{2015}\natexlab{}.
\newblock \showarticletitle{Semantic path based personalized recommendation on
  weighted heterogeneous information networks}. In
  \bibinfo{booktitle}{\emph{CIKM}}.
\newblock


\bibitem[\protect\citeauthoryear{Tintarev and Masthoff}{Tintarev and
  Masthoff}{2007}]%
        {tintarev2007survey}
\bibfield{author}{\bibinfo{person}{Nava Tintarev} {and} \bibinfo{person}{Judith
  Masthoff}.} \bibinfo{year}{2007}\natexlab{}.
\newblock \showarticletitle{A survey of explanations in recommender systems}.
  In \bibinfo{booktitle}{\emph{Workshop on Ambient Intelligence, Media and
  Sensing}}.
\newblock


\bibitem[\protect\citeauthoryear{Wan and McAuley}{Wan and McAuley}{2018}]%
        {wan2018item}
\bibfield{author}{\bibinfo{person}{Mengting Wan} {and} \bibinfo{person}{Julian
  McAuley}.} \bibinfo{year}{2018}\natexlab{}.
\newblock \showarticletitle{Item recommendation on monotonic behavior chains}.
  In \bibinfo{booktitle}{\emph{RecSys}}.
\newblock


\bibitem[\protect\citeauthoryear{Wang, Zhang, Wang, Zhao, Li, Xie, and
  Guo}{Wang et~al\mbox{.}}{2018c}]%
        {wang2018ripplenet}
\bibfield{author}{\bibinfo{person}{Hongwei Wang}, \bibinfo{person}{Fuzheng
  Zhang}, \bibinfo{person}{Jialin Wang}, \bibinfo{person}{Miao Zhao},
  \bibinfo{person}{Wenjie Li}, \bibinfo{person}{Xing Xie}, {and}
  \bibinfo{person}{Minyi Guo}.} \bibinfo{year}{2018}\natexlab{c}.
\newblock \showarticletitle{{Ripplenet: Propagating user preferences on the
  knowledge graph for recommender systems}}. In
  \bibinfo{booktitle}{\emph{CIKM}}.
\newblock


\bibitem[\protect\citeauthoryear{Wang, Wang, Jia, and Yin}{Wang
  et~al\mbox{.}}{2018b}]%
        {wang2018explainable}
\bibfield{author}{\bibinfo{person}{Nan Wang}, \bibinfo{person}{Hongning Wang},
  \bibinfo{person}{Yiling Jia}, {and} \bibinfo{person}{Yue Yin}.}
  \bibinfo{year}{2018}\natexlab{b}.
\newblock \showarticletitle{Explainable recommendation via multi-task learning
  in opinionated text data}. In \bibinfo{booktitle}{\emph{SIGIR}}.
\newblock


\bibitem[\protect\citeauthoryear{Wang, Chen, Yang, Wu, Wu, and Xie}{Wang
  et~al\mbox{.}}{2018a}]%
        {wang2018reinforcement}
\bibfield{author}{\bibinfo{person}{Xiting Wang}, \bibinfo{person}{Yiru Chen},
  \bibinfo{person}{Jie Yang}, \bibinfo{person}{Le Wu},
  \bibinfo{person}{Zhengtao Wu}, {and} \bibinfo{person}{Xing Xie}.}
  \bibinfo{year}{2018}\natexlab{a}.
\newblock \showarticletitle{A Reinforcement Learning Framework for Explainable
  Recommendation}. In \bibinfo{booktitle}{\emph{ICDM}}.
\newblock


\bibitem[\protect\citeauthoryear{Wang, Wang, Xu, He, Cao, and Chua}{Wang
  et~al\mbox{.}}{2019}]%
        {wang2019explainable}
\bibfield{author}{\bibinfo{person}{Xiang Wang}, \bibinfo{person}{Dingxian
  Wang}, \bibinfo{person}{Canran Xu}, \bibinfo{person}{Xiangnan He},
  \bibinfo{person}{Yixin Cao}, {and} \bibinfo{person}{Tat-Seng Chua}.}
  \bibinfo{year}{2019}\natexlab{}.
\newblock \showarticletitle{Explainable reasoning over knowledge graphs for
  recommendation}. In \bibinfo{booktitle}{\emph{AAAI}}.
\newblock


\bibitem[\protect\citeauthoryear{Xian, Fu, Muthukrishnan, de~Melo, and
  Zhang}{Xian et~al\mbox{.}}{2019}]%
        {xian2019reinforcement}
\bibfield{author}{\bibinfo{person}{Yikun Xian}, \bibinfo{person}{Zuohui Fu},
  \bibinfo{person}{S. Muthukrishnan}, \bibinfo{person}{Gerard de Melo}, {and}
  \bibinfo{person}{Yongfeng Zhang}.} \bibinfo{year}{2019}\natexlab{}.
\newblock \showarticletitle{Reinforcement Knowledge Graph Reasoning for
  Explainable Recommendation}. In \bibinfo{booktitle}{\emph{SIGIR}}.
\newblock


\bibitem[\protect\citeauthoryear{Yang, Liu, Wang, and Hu}{Yang
  et~al\mbox{.}}{2018}]%
        {DBLP:conf/icdm/YangLWH18}
\bibfield{author}{\bibinfo{person}{Fan Yang}, \bibinfo{person}{Ninghao Liu},
  \bibinfo{person}{Suhang Wang}, {and} \bibinfo{person}{Xia Hu}.}
  \bibinfo{year}{2018}\natexlab{}.
\newblock \showarticletitle{Towards Interpretation of Recommender Systems with
  Sorted Explanation Paths}. In \bibinfo{booktitle}{\emph{ICDM}}.
\newblock


\bibitem[\protect\citeauthoryear{Yu, Ren, Sun, Gu, Sturt, Khandelwal, Norick,
  and Han}{Yu et~al\mbox{.}}{2014}]%
        {yu2014personalized}
\bibfield{author}{\bibinfo{person}{Xiao Yu}, \bibinfo{person}{Xiang Ren},
  \bibinfo{person}{Yizhou Sun}, \bibinfo{person}{Quanquan Gu},
  \bibinfo{person}{Bradley Sturt}, \bibinfo{person}{Urvashi Khandelwal},
  \bibinfo{person}{Brandon Norick}, {and} \bibinfo{person}{Jiawei Han}.}
  \bibinfo{year}{2014}\natexlab{}.
\newblock \showarticletitle{{Personalized entity recommendation: A
  heterogeneous information network approach}}. In
  \bibinfo{booktitle}{\emph{WSDM}}.
\newblock


\bibitem[\protect\citeauthoryear{Yu, Ren, Sun, Sturt, Khandelwal, Gu, Norick,
  and Han}{Yu et~al\mbox{.}}{2013}]%
        {yu2013recommendation}
\bibfield{author}{\bibinfo{person}{Xiao Yu}, \bibinfo{person}{Xiang Ren},
  \bibinfo{person}{Yizhou Sun}, \bibinfo{person}{Bradley Sturt},
  \bibinfo{person}{Urvashi Khandelwal}, \bibinfo{person}{Quanquan Gu},
  \bibinfo{person}{Brandon Norick}, {and} \bibinfo{person}{Jiawei Han}.}
  \bibinfo{year}{2013}\natexlab{}.
\newblock \showarticletitle{Recommendation in heterogeneous information
  networks with implicit user feedback}. In \bibinfo{booktitle}{\emph{RecSys}}.
\newblock


\bibitem[\protect\citeauthoryear{Zhang, Swami, and Chawla}{Zhang
  et~al\mbox{.}}{2019b}]%
        {zhang2019shne}
\bibfield{author}{\bibinfo{person}{Chuxu Zhang}, \bibinfo{person}{Ananthram
  Swami}, {and} \bibinfo{person}{Nitesh Chawla}.}
  \bibinfo{year}{2019}\natexlab{b}.
\newblock \showarticletitle{{SHNE: Representation Learning for
  Semantic-Associated Heterogeneous Networks}}. In
  \bibinfo{booktitle}{\emph{WSDM}}.
\newblock


\bibitem[\protect\citeauthoryear{Zhang, Lofgren, and Goel}{Zhang
  et~al\mbox{.}}{2016}]%
        {zhang2016approximate}
\bibfield{author}{\bibinfo{person}{Hongyang Zhang}, \bibinfo{person}{Peter
  Lofgren}, {and} \bibinfo{person}{Ashish Goel}.}
  \bibinfo{year}{2016}\natexlab{}.
\newblock \showarticletitle{Approximate personalized pagerank on dynamic
  graphs}. In \bibinfo{booktitle}{\emph{KDD}}.
\newblock


\bibitem[\protect\citeauthoryear{Zhang, Fan, Zhou, Todorovic, Wu, and Wu}{Zhang
  et~al\mbox{.}}{2019a}]%
        {zhang2019cvpr}
\bibfield{author}{\bibinfo{person}{Quanshi Zhang}, \bibinfo{person}{Lixin Fan},
  \bibinfo{person}{Bolei Zhou}, \bibinfo{person}{Sinisa Todorovic},
  \bibinfo{person}{Tianfu Wu}, {and} \bibinfo{person}{Ying~Nian Wu}.}
  \bibinfo{year}{2019}\natexlab{a}.
\newblock \showarticletitle{{CVPR-19 Workshop on Explainable AI}}.
\newblock


\bibitem[\protect\citeauthoryear{Zhang and Chen}{Zhang and Chen}{2018}]%
        {zhang2018explainable}
\bibfield{author}{\bibinfo{person}{Yongfeng Zhang} {and} \bibinfo{person}{Xu
  Chen}.} \bibinfo{year}{2018}\natexlab{}.
\newblock \showarticletitle{{Explainable recommendation: A} survey and new
  perspectives}.
\newblock \bibinfo{journal}{\emph{arXiv preprint arXiv:1804.11192}}
  (\bibinfo{year}{2018}).
\newblock


\bibitem[\protect\citeauthoryear{Zhang, Lai, Zhang, Zhang, Liu, and Ma}{Zhang
  et~al\mbox{.}}{2014}]%
        {zhang2014explicit}
\bibfield{author}{\bibinfo{person}{Yongfeng Zhang}, \bibinfo{person}{Guokun
  Lai}, \bibinfo{person}{Min Zhang}, \bibinfo{person}{Yi Zhang},
  \bibinfo{person}{Yiqun Liu}, {and} \bibinfo{person}{Shaoping Ma}.}
  \bibinfo{year}{2014}\natexlab{}.
\newblock \showarticletitle{Explicit factor models for explainable
  recommendation based on phrase-level sentiment analysis}. In
  \bibinfo{booktitle}{\emph{SIGIR}}.
\newblock


\bibitem[\protect\citeauthoryear{Zhang, Zhang, Zhang, and Shah}{Zhang
  et~al\mbox{.}}{2019c}]%
        {zhang2019ears}
\bibfield{author}{\bibinfo{person}{Yongfeng Zhang}, \bibinfo{person}{Yi Zhang},
  \bibinfo{person}{Min Zhang}, {and} \bibinfo{person}{Chirag Shah}.}
  \bibinfo{year}{2019}\natexlab{c}.
\newblock \showarticletitle{{EARS 2019: The 2nd International Workshop on
  ExplainAble Recommendation and Search}}. In
  \bibinfo{booktitle}{\emph{SIGIR}}.
\newblock


\end{thebibliography}

\end{document}